\documentclass[10pt,twocolumn,letterpaper]{article}

\usepackage[pagenumbers]{cvpr} % To force page numbers, e.g. for an arXiv version

\usepackage[T1]{fontenc}

\definecolor{myblue}{rgb}{0.21,0.49,0.74}
\usepackage[pagebackref,breaklinks,colorlinks,allcolors=myblue]{hyperref}

\definecolor{ourgreen}{HTML}{d2f0aa}
\definecolor{ouryellow}{HTML}{fcef9e}

\title{Mask-Free Audio-driven Talking Face Generation for Enhanced Visual Quality and Identity Preservation}

\author{Dogucan Yaman\textsuperscript{1} \qquad Fevziye Irem Eyiokur\textsuperscript{1} \qquad Leonard Bärmann\textsuperscript{1} \\ Hazım Kemal Ekenel\textsuperscript{2} \qquad Alexander Waibel\textsuperscript{1,3} \\
\textsuperscript{1}Karlsruhe Institute of Technology, \textsuperscript{2}Istanbul Technical University, \textsuperscript{3}Carnegie Mellon University\\
{\tt\small dogucan.yaman@kit.edu}}

\begin{document}
\maketitle
\begin{abstract}
Audio-Driven Talking Face Generation aims at generating realistic videos of talking faces, focusing on accurate audio-lip synchronization without deteriorating any identity-related visual details.
Recent state-of-the-art methods are based on inpainting, meaning that the lower half of the input face is masked, and the model fills the masked region by generating lips aligned with the given audio.
Hence, to preserve identity-related visual details from the lower half, these approaches additionally require an unmasked identity reference image randomly selected from the same video. 
However, this common masking strategy suffers from
(1) information loss in the input faces, significantly affecting the networks' ability to preserve visual quality and identity details,
(2) variation between identity reference and input image degrading reconstruction performance, and
(3) the identity reference negatively impacting the model, causing unintended copying of elements unaligned with the audio.
To address these issues, we propose a mask-free talking face generation approach while maintaining the 2D-based face editing task.
Instead of masking the lower half, we transform the input images to have closed mouths, using a two-step landmark-based approach trained in an unpaired manner.
Subsequently, we provide these edited but unmasked faces to a lip adaptation model alongside the audio to generate appropriate lip movements.
Thus, our approach needs neither masked input images nor identity reference images.
We conduct experiments on the benchmark LRS2 and HDTF datasets and perform various ablation studies to validate our contributions.
\end{abstract}
    
\section{Introduction}
\label{sec:intro}

\begin{figure}[tb]
  \centering
  \begin{subfigure}{\linewidth}
    \includegraphics[width=\linewidth]{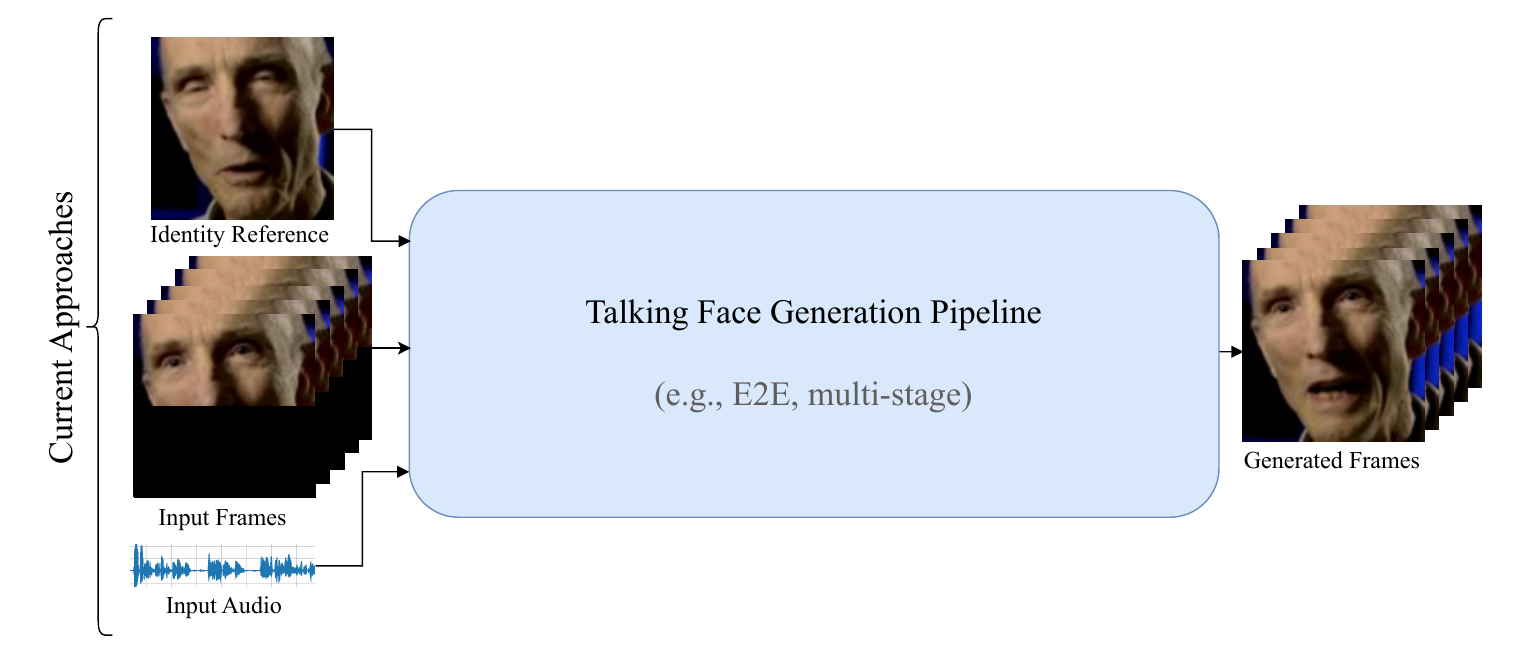}
    \caption{Traditional talking face generation pipeline.}
    \label{fig:overview:traditional}
  \end{subfigure}
  
  \begin{subfigure}{\linewidth}
    \includegraphics[trim={0cm 0cm 0cm 0cm},clip,width=\linewidth]{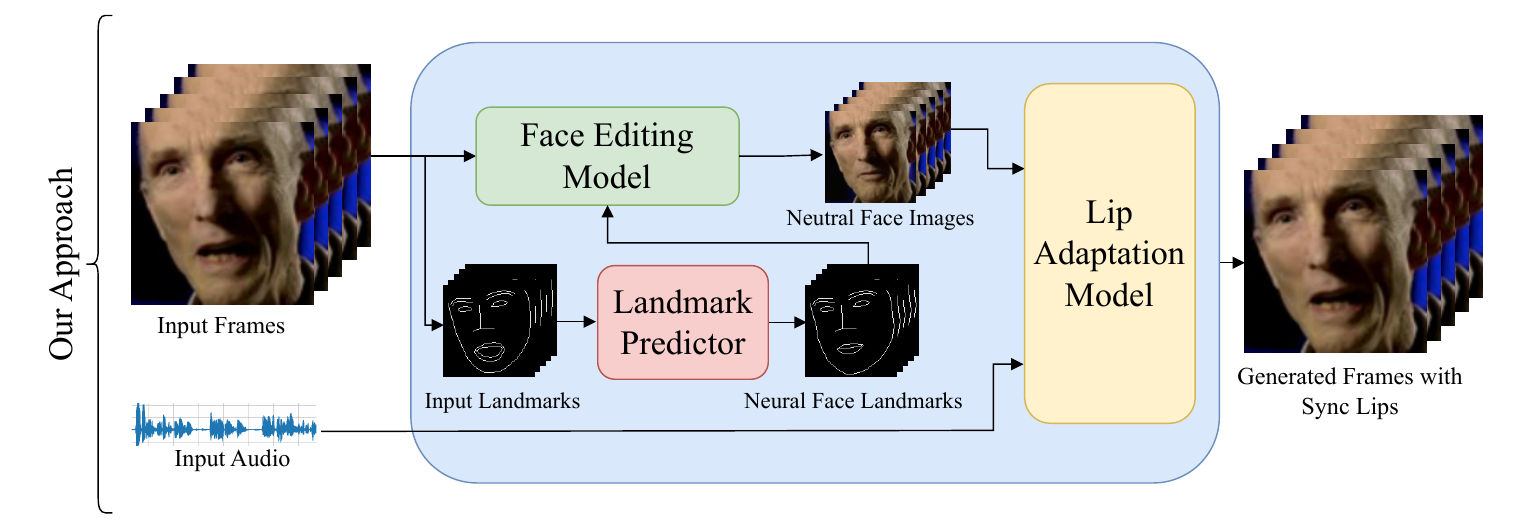}
    \caption{Our mask-free approach.}
    \label{fig:overview:mask_free}
  \end{subfigure}
  
  \caption{Demonstration of the traditional talking face generation approach and our mask-free approach.}
  \label{fig:overview}
\end{figure}

Audio-driven 2D talking face generation, a.k.a. lip reanimation, generates a video by manipulating the lips of existing video frames with respect to a given audio, while preserving visual and identity-related details.
Talking face generation has gained significant popularity due to its potential in applications like virtual assistants, video/movie dubbing, and digital content creation \& translation~\cite{zhan2023multimodal,zhen2023human,jiang2024audio}.
In this intricate task, lip-sync and visual quality are essential factors for generating natural-looking videos.
While lip-sync ensures that lip movements are aligned with the audio, visual quality involves delivering high-resolution, artifact-free visual content that also preserves the subject's identity.
Any issues in these details make the video less natural since they are easily recognizable by humans.

\begin{figure}[tb]
    \centering
    \includegraphics[trim={0.8cm 0.5cm 0.7cm 0.8cm},clip,width=\linewidth]{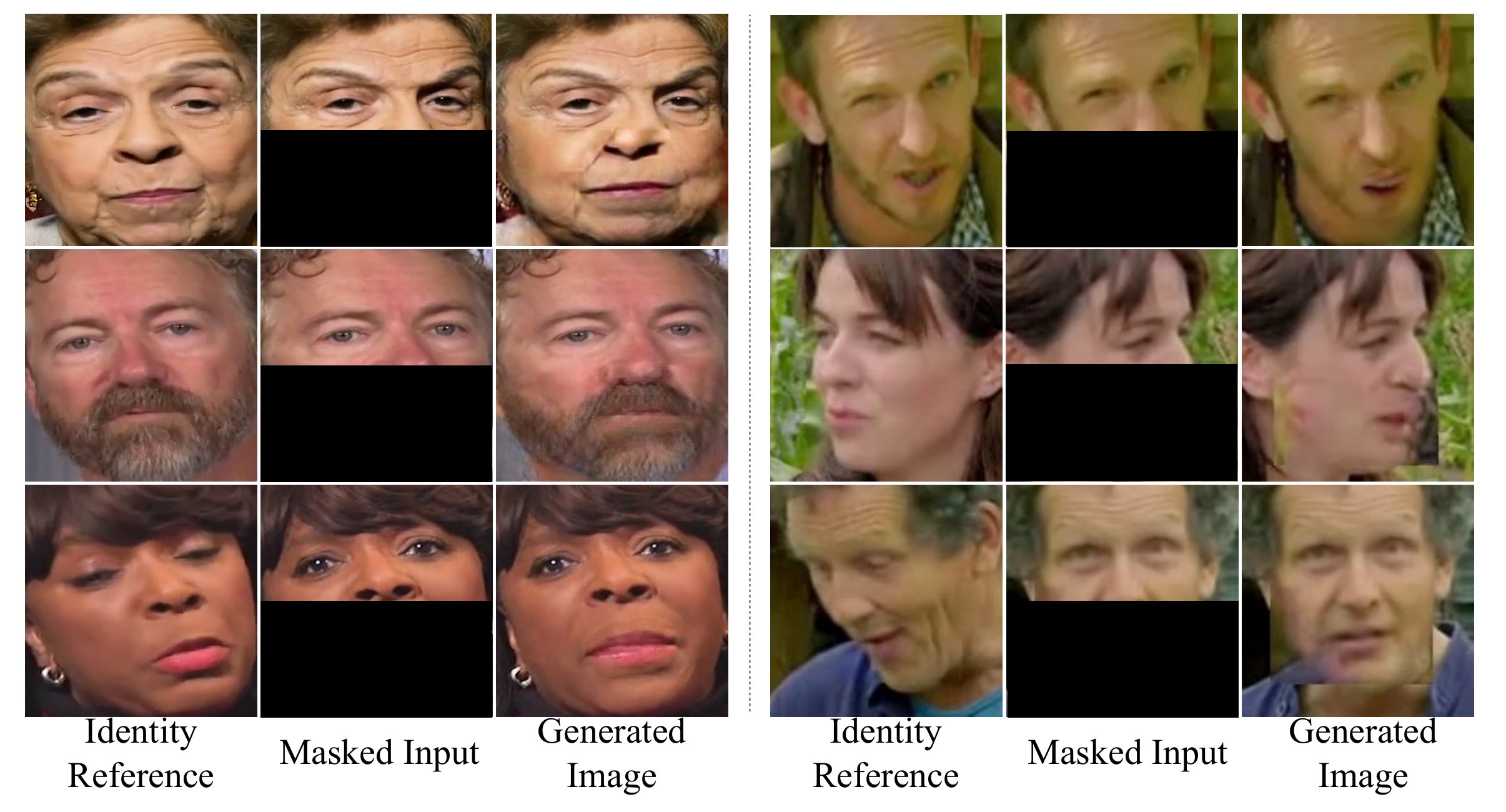}
    \caption{Mask-related problems. Generated images are clearly influenced by the identity reference. Further, masking leads to occasional errors in details of pose, background, face borders etc. Images generated by: left: our baseline experiments, right (top to bottom): \cite{cheng2022videoretalking,wang2023seeing,prajwal2020lip}. Best viewed by zooming in.}
    \label{fig:issues_all}
\end{figure}

To achieve these goals, recent approaches in the literature \cite{prajwal2020lip,shen2023difftalk,zhong2023identity} use an inpainting-based scheme (\cref{fig:overview:traditional}):
A generative network receives the input audio and a sequence of video frames with masked mouth region concealing the ground-truth lip shape (e.g. by masking the lower half of the face), and is trained to reconstruct the masked part aligned with the given audio.
This is done using a combination of loss functions including simple reconstruction loss, adversarial loss, and specialized lip-sync loss utilizing pretrained feature extraction networks to measure audio-lip synchronization \cite{prajwal2020lip}.
However, the described masking necessarily leads to information loss, meaning that important identity-related details would be missing.
Therefore, the model additionally receives one (or multiple) identity reference image(s), typically selected randomly from different time steps of the input video.

This straightforward approach, nevertheless, has fundamental drawbacks:
(1) The masking strategy causes a loss of available information and requires the network to regenerate the entire masked region using the available information from the identity reference and the upper part of the face.
This hardens the network's task and sometimes hinders accurate inference of the missing details and preservation of identity, despite the identity reference image.
Generally, predicting more content of the image raises the likelihood of errors and artifacts.
(2) The differences between the identity reference and masked input image in lightning, pose, and expression can complicate the reconstruction process, resulting in visual artifacts and alignment problems.

(3) The identity reference can undesirably influence the model, leading to issues like lip leaking~\cite{prajwal2020lip,cheng2022videoretalking,yaman2024audiodriventalkingfacegeneration}, where the model occasionally copies the lip shape of the identity reference although it is unaligned with the audio, both in training and inference.
Thus, the model ends up with suboptimal lip-sync and visual quality~\cite{prajwal2020lip,cheng2022videoretalking,yaman2024audiodriventalkingfacegeneration}, as illustrated in \cref{fig:issues_all}. 

In this work, we circumvent these issues by introducing a mask-free talking face generation approach (MF-Talk, \cref{fig:overview:mask_free}).
On a high level, instead of masking the faces in the input video frames, we transform them to always have closed lips.
Given such a sequence of closed-mouth frames and the input audio, our model generates faces aligned with the audio, without requiring an additional identity reference since the input images are not masked.
Specifically, we begin by training a transformer-based lip landmark prediction model responsible for generating lip landmarks that accurately represent closed and flat mouths, i.e., silent lips~\cite{cheng2022videoretalking,yaman2024audiodriventalkingfacegeneration}.
Next, we train a landmark-driven face editing model in an unpaired manner to modify the lips of the input image to appear closed, using the predicted landmarks as a condition.
Finally, we use the modified image sequence as input to our lip adaptation model, along with audio, to generate a face sequence by only editing the lips, neither using masking nor an identity reference.
With our mask-free approach, we can benefit from the existing information in the input image and simplify the task by editing only a small portion of the input rather than generating the masked region from scratch by trying to acquire the missing information from the identity reference.
Our contributions are:
\begin{itemize}
    \item We introduce mask-free talking face generation (MF-Talk) for the first time, as an alternative to the inpainting-based approaches, more accurately preserving identity, improving visual quality, and simplifying the model's learning problem (see \cref{fig:overview}). %Besides, our talking face generation model does not require any reference image.
    \item Our approach is able to synthesize the video with the appropriate lip movements by only using the input face sequence, without requiring an identity reference image, thus alleviating many issues of existing methods.
    \item We propose a landmark prediction model that accurately generates landmarks to represent neutral/closed mouth and a face editing model conditioned on predicted landmark maps for face generation with neutral/closed mouth.% (face with closed lips / mouth). % and train it in an unpaired manner.
    \item We conduct extensive experiments and detailed analyses to show the effectiveness of our approach.
    
\end{itemize}

\section{Related Work}
\label{sec:related_work}

\subsection{Masking Strategy}

Early research efforts use mapping between audio features and time-aligned facial motions~\cite{yehia1998quantitative} and perform facial motion prediction using HMMs~\cite{brand1999voice}.
A more recent study~\cite{suwajanakorn2017synthesizing} generates video by retrieving the images that are most aligned with the audio.
Another approach to talking face generation is to use facial landmark representations and generate the video based on these, as directly mapping audio to face is more challenging~\cite{chen2019hierarchical,das2020speech,zhou2020makelttalk,zhong2023identity}.
One of the major milestones in talking face generation is Wav2Lip~\cite{prajwal2020lip}, which addresses the task as an audio-conditioned image inpainting task.
For this, the faces in a video are processed as a sequence in each step.
The lower half of the faces is masked and provided to the image encoder along with a randomly selected identity reference, since the identity-related details in the input faces are unavailable due to the masking strategy.
This approach demonstrates superior performance in both lip-sync accuracy and identity preservation.
Due to its effectiveness, subsequent works apply the same masking strategy to treat the task as image inpainting~\cite{park2022synctalkface,cheng2022videoretalking,wang2023lipformer,zhang2023dinet,zhong2023identity,wang2023seeing,muaz2023sidgan,shen2023difftalk,stypulkowski2024diffused,ma2023dreamtalk,yaman2024audio,yaman2024audiodriventalkingfacegeneration,peng2024synctalk,mukhopadhyay2024diff2lip}.
In contrast to this, we propose a mask-free approach, by first transforming the video frames to have closed lips, and subsequently using these frames for audio-driven talking face generation.
This way, we circumvent various issues that arise from masking and the necessity of an identity reference in the inpainting-based methods.

\subsection{Lip-sync Learning}

Lip-sync learning is a central point of the talking face generation task.
While initial studies utilize hand-crafted features and statistical models~\cite{hershey1999audio,slaney2000facesync}, later approaches focus on benefiting from mutual information between audio-visual features to predict output as sync or not sync for sound~\cite{chen2021audio,iashin2022sparse,owens2018audio} and speech~\cite{afouras2020self,chung2017lip,chung2019perfect,kadandale2022vocalist,kim2021end}.
Although some works learn lip-sync implicitly~\cite{chen2019hierarchical,guo2021ad,suwajanakorn2017synthesizing,wu2021imitating,zhang2021flow,jamaludin2019you,kr2019towards,tian2024emo}, 
explicit learning improves lip-sync accuracy, especially with limited data.
Some methods~\cite{ji2021audio,papantoniou2022neural,song2022everybody,zhou2020makelttalk,zhong2023identity} employ landmark distance to guide the model in lip-sync learning.
However, they lack optimal lip-sync despite good visual quality and stability.
So far, the most common and accurate method for lip-sync learning is to employ an additional network for multimodal feature extraction and to compute a loss representing the synchronization between the generated lip movements and the given audio snippet~\cite{liang2022expressive,park2022synctalkface,prajwal2020lip,wang2023progressive,wang2023lipformer,wang2023seeing,yao2022dfa,zhang2023dinet,zhou2019talking,zhou2021pose,eskimez2020end,guan2023stylesync,song2018talking,sun2022masked,vougioukas2020realistic,yaman2024audio,yaman2024audiodriventalkingfacegeneration}.
In this work, we follow recently proposed stabilized synchronization loss~\cite{yaman2024audiodriventalkingfacegeneration}, that alleviates lip-sync learning-related issues, and adapt it for our approach.

\subsection{Portrait Animation}

Rather than solely editing faces in 2D for video translation, portrait animation (a.k.a talking head generation) involves generating an entire video from either a single image (one-shot) or using all the frames for extracting corresponding parameters (e.g., pose, identity, expression) and regenerating entire head. 
While some methods perform this head generation in 2D space~\cite{zhou2021pose,liang2022expressive,tian2024emo}, the majority of works prefer 3D-based methods and Neural Radian Fields (NeRFs) for more precise control in generation~\cite{blanz2023morphable,booth2018large,guo2021ad,liu2022semantic,papantoniou2022neural,shen2022learning,song2022everybody,tang2022real,thies2020neural,wang2023progressive,wu2021imitating,yao2022dfa,ye2023geneface,yin2022styleheat,zhang2021facial,zhang2021flow,zhou2019talking,zhou2020makelttalk,zhang2023sadtalker,kim2024nerffacespeech,zhang2024learning,chu2024gpavatar,ma2024cvthead,li2023efficient,liu2023moda,chen2023implicit,wu2023ganhead,li2023one,liu2023font,ye2023geneface,ma2023otavatar}.
In portrait animation, expression and pose controllability~\cite{ji2021audio,zhou2021pose,liang2022expressive,ji2022eamm,tan2023emmn,gan2023efficient,zhang2024emotalker,zhang2024emodiffhead,chakkera2024jean,liu2023opt,xu2023high,sun2023vividtalk,hwang2023discohead} are essential for creating a natural video, as all parameters are individually available.
However, this is quite challenging. Therefore, 2D editing-based approaches are necessary when preserving the details of the video is crucial, such as in movie dubbing.
Similarly, some works go further than just head generation,
exploring talking head video generation that includes natural torso~\cite{ye2024real3d} and full-body gestures~\cite{hogue2024diffted,chatziagapi2024talkinnerf}.
On the other hand, transferring or controlling a speaking style is another research direction in the literature~\cite{ma2023styletalk,song2022audio,wang2024styletalk++}.
Please note that the task that we cover in this paper involves frame-by-frame 2D video editing to achieve precise lip synchronization with audio, making it a fundamentally different approach to portrait animation.

\section{Mask-Free Talking Face Generation}

\begin{figure*}[t]
    \centering
    \includegraphics[width=\linewidth]{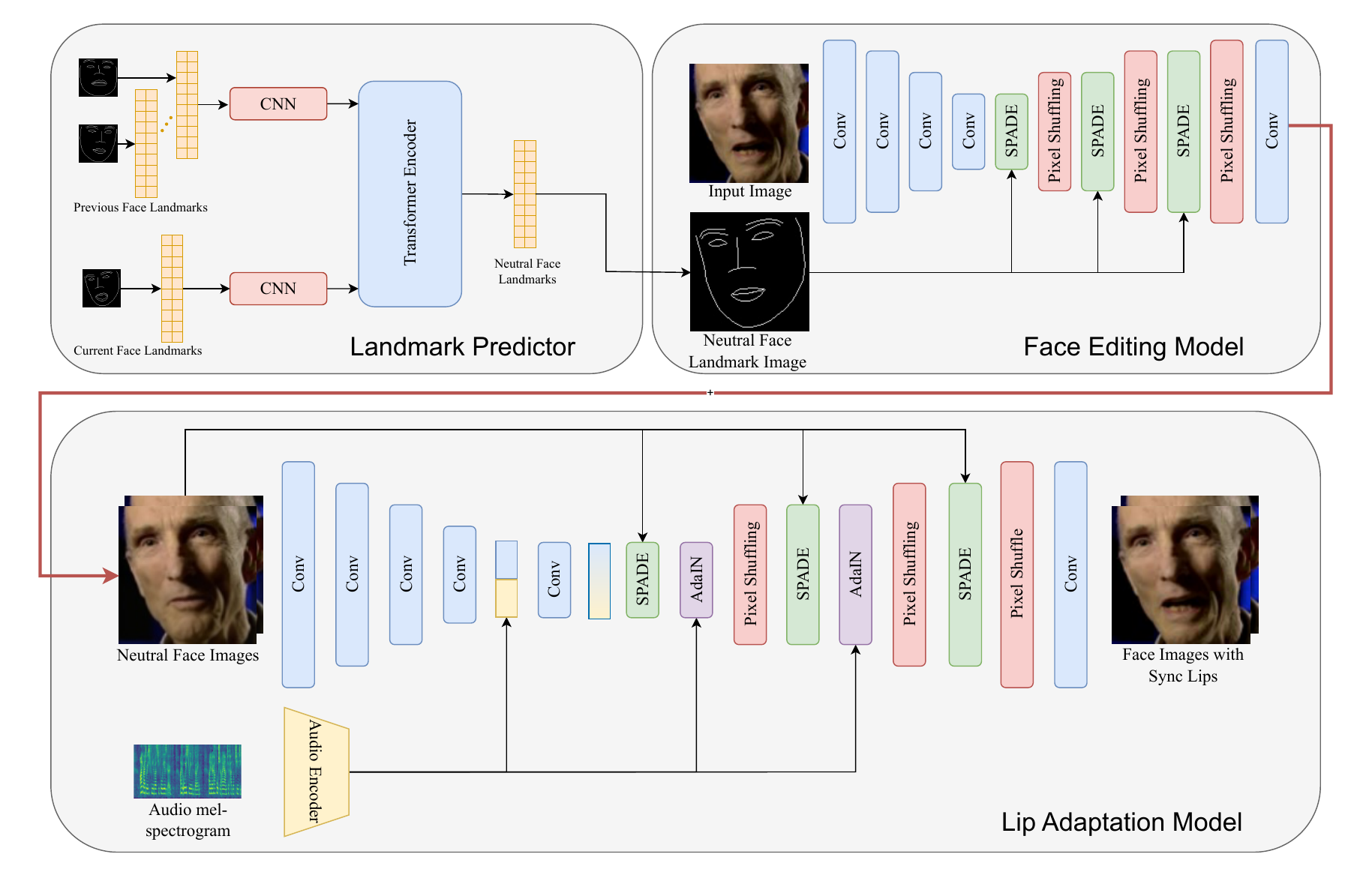}
    \caption{Our mask-free talking face generation pipeline in inference. First, a landmark predictor ($T_L$) generates landmarks for neutral mouth. Next, our face editing model ($G_E$) utilizes the generated neutral face landmarks to modify the input image to have a neutral mouth. Finally, the lip adaptation model ($G_L$) employs the output of the face editing model along with the audio input to generate sync lips.}
    \label{fig:method}
\end{figure*}

In this paper, we propose mask-free talking face generation (MF-Talk), aiming to better preserve the identity, removing mask-related artifacts, and eliminating negative influences of the identity reference.
As shown in \cref{fig:method}, we decompose the problem into three subtasks: neutral landmark prediction, landmark-driven face editing for neutral mouth generation, and lip adaptation.

In our method, we first extract a landmark map for each face using Mediapipe~\cite{lugaresi2019mediapipe}. 
Then, we predict the new neutral-mouth landmarks using our transformer-based landmark prediction model ($T_L$).
Next, our landmark-driven face editing model ($G_E$) takes the input image and the predicted landmark map to modify the mouth region of the image.
Finally, the lip adaptation model ($G_L$) gets the audio input and the modified neutral-mouth input face to adapt the mouth region, generating proper lip movements with respect to the given audio to achieve synchronized lips.

\subsection{Landmark Prediction Model}
 
We first extract face landmark vectors $l_t \in \mathbb{R}^{2 \times 131}$ from each video frame at timestep $t$.
Given a such a vector, we write $l_t = (l^l_t, l^j_t, l^p_t)$ with lip landmarks $l^l_t \in \mathbb{R}^{2 \times 41}$, jaw landmarks $l^j_t \in \mathbb{R}^{2 \times 16}$, and pose landmarks $l^p_t \in \mathbb{R}^{2 \times 74}$.
Our goal is to train a transformer-based model $ T_L $ to predict lip and jaw landmarks given all landmarks from the $k$ previous frames and the current frame's pose landmarks:%, specifically:
\begin{equation}
    T_L : \left(\,l_{t-k}\,,\,...\,,\,l_{t-1}\,,\, l^p_t\,\right) \to (\,l^l_t\,,\,l^j_t\,)
\end{equation}
Since we aim to predict lip and jaw landmarks, we provide the remaining landmarks from the current frame to the network, which we refer to as pose landmarks.
Moreover, we provide the landmarks from previous frames to guide the model in making smooth predictions and obtain identity information.
$T_L$ is responsible for predicting landmarks to represent neutral mouth (see \Cref{sec:method:training} for training details) as these landmarks are used in $G_E$ as a condition. 
Note that we represent all landmarks in $T_L$ as normalized coordinate pairs. The sketch inputs in Fig. \ref{fig:method} are provided for illustration purposes only.

Our $ T_L $ consists of two parallel encoders with 1D convolutions to encode the landmarks of the previous frames, $ E_r $, and pose landmarks of the current frame, $ E_p $, followed by a transformer encoder with four layers. 
Each layer includes multi-head self-attention (MHSA)~\cite{vaswani2017attention}, layer normalization (LN)~\cite{lei2016layer}, and multi-layer perceptron layers (MLP).
\begin{equation}
    F^{l^p_t} = E_p(l^p_t)
\end{equation}
\begin{equation}
    F^{l_{t-1}} = E_r(l_{t-1})
\end{equation}
\begin{equation}
    F^{l}_i = \text{TE}_i(F^{l}_{i-1})
\end{equation}
where $ \text{TE}_i $ indicates layer $i$ of the transformer encoder. 
We apply MLP on top of the transformer’s hidden states to generate the output prediction $\hat{l}_t^l, \hat{l}_t^j$.
To train $ T_L $, we utilize L1 reconstruction loss between GT landmarks and generated landmarks, i.e. the landmark reconstruction loss:
\begin{equation}
    L_l = ||\hat{l}_t^j - l_t^j|| + \lambda ||\hat{l}_t^l - l_t^l||
\end{equation}
where $ \lambda $ is set as $ 10 $ to focus on the lip landmarks more.
Since the GT data is a subset of LRS2 dataset, which consists of closed-mouth samples only, our $T_L$ can learn to generate a landmark map with a closed mouth from any input landmark map while preserving both pose and identity.

\subsection{Landmark-driven Face Editing Model }
\label{sec:face_editing_model}
We utilize a GAN-based~\cite{goodfellow2014generative} conditional image editing model ($G_E$) that takes an input image and the landmark map to edit the mouth region of the face, synthesizing the same image with neutral mouth that matches the input landmark map.
Our $ G_E $ has a U-Net shape architecture~\cite{ronneberger2015u}, contains an image encoder and an image decoder with SPADE~\cite{park2019semantic} and Pixel Shuffling~\cite{shi2016real} layers (see \cref{app:landmark_driven_face_editing_model} for details).
To train $ G_E $, we employ adversarial loss~\cite{goodfellow2014generative}, perceptual loss~\cite{johnson2016perceptual} with pretrained VGG-19~\cite{simonyan2014very} features, feature matching loss~\cite{wang2018high}, and landmark reconstruction loss to match the lip and jaw landmarks of the generated image with the input landmarks.
Moreover, we utilize an additional pretrained model that focuses on only the mouth region to classify it as open or closed (see \Cref{app:mouth_classification_model}).
We train this model by labeling the training images in LRS2 dataset~\cite{LRS2} as open / closed mouth according to the distance between the landmarks of the upper and lower lips.
The training objective of our $G_E$ is as follows:
\begin{equation}
    L_E = L_{GAN} + \lambda_1 L_{per} + \lambda_2 L_{FM} + \lambda_3 L_{l} + L_{m}
\end{equation}
where $L_{per}$ indicates perceptual loss, $L_{FM}$ states feature matching loss, $L_{l}$ represents landmark reconstruction loss, and $L_{m}$ is cross-entropy loss for mouth classification model, which works like a discriminator.
Since we don't mask the input image and employ $L_{per}$ along with $L_{FM}$, our $G_E$ effectively preserves identity while editing the lips (see Table \ref{tab:silent_models}).
($\lambda_1, \lambda_2, \lambda_3$) ) ($1, 0.1, 0.25$).

\begin{table*}[!ht]
  \centering
  \resizebox{\textwidth}{!}{\begin{tabular}{@{}l|ccccccc | ccccccc@{}}
    \toprule
     & \multicolumn{7}{c}{\textbf{LRS2}} & \multicolumn{7}{c}{\textbf{HDTF}} \\
    \midrule
    Method & SSIM $\uparrow$ & PSNR $\uparrow$ & FID $\downarrow$ & LMD $\downarrow$ & LSE-C $\uparrow$ & LSE-D $\downarrow$ & CSIM $\uparrow$ & SSIM $\uparrow$ & PSNR $\uparrow$ & FID $\downarrow$ & LMD $\downarrow$ & LSE-C $\uparrow$ & LSE-D $\downarrow$ & CSIM $\uparrow$ \\
    \hline
    Wav2Lip~\cite{prajwal2020lip} & 0.86 & 26.53 & 7.05 & 2.38 & 7.59 & 6.75 & {0.84} & 0.84 & 24.81 & 35.41 & {1.34} & {9.05} & {6.14} & 0.87 \\
    %PC-AVS~\cite{zhou2021pose} & 0.73 & 28.24  & 18.40 & 1.93 & 6.41 & 7.52 &  &  &  &  &  &   \\
    %EAMM~\cite{ji2022eamm} & 0.69 & 21.01  & 84.65 & 0.51 & 3.54 & 3.31 & 9.93 & 0.71 & 26.22 & 44.16 & 0.48 & 2.61 & 4.32 & 9.04 \\
    VideoReTalking w/ FR~\cite{cheng2022videoretalking} & 0.84 & 25.58 & 9.28 & 2.61 & 7.49 & 6.82 & 0.75 & 0.83 & 24.55 & 29.77 & 3.09 & 6.12 & 7.37 & {0.89} \\
    DINet~\cite{zhang2023dinet} & 0.78 & 24.35 & {4.26} & 2.30 & 5.37 & 8.37 & 0.73 & 0.91 & 29.12 & 18.77 & 1.45 & 6.42 & 8.93 & 0.82 \\ % 0.9085
    TalkLip~\cite{wang2023seeing} & 0.86 & 26.11  & 4.94 & 2.34 & {8.53} & {6.08} & 0.75 & 0.82 & 25.23 & 25.10 & 2.98 & 6.19 & {7.28} & 0.89 \\
    IPLAP~\cite{zhong2023identity} & {0.87} & {29.67} & {4.10} &{2.11} & 6.49 & 7.16 & 0.82 & {0.87} & {27.80} & {22.09} & 2.21 & 5.56  & 8.49 & 0.80 \\
    AVTFG~\cite{yaman2024audio} & 0.95 & 31.27 & 4.51 & 1.19 & 7.95 & 6.30 & 0.80 & {0.93} & {30.58} & {16.76} & {1.29} & {8.11} & {6.77} & 0.89 \\
    PLGAN~\cite{yaman2024audiodriventalkingfacegeneration} & {0.95} & {32.64} & {3.83} & {1.13} & {8.41} & {6.03} & 0.79 & 0.89 & 28.60 & 21.46 & {1.30} & {8.30} & {6.36} & 0.81  \\
    Diff2Lip~\cite{mukhopadhyay2024diff2lip} & 0.94 & 31.68 & 3.80 & 1.50 & 7.87 & 6.46 & 0.85 & 0.83 & 26.07 & 27.82 & 2.29 & 7.45 & 7.16 & 0.81 \\ % correct one
    \hline
    Ours & \textbf{0.95} & \textbf{33.96}  & \textbf{3.57} & {1.18} & {7.76} & {6.32} & \textbf{0.88} & \textbf{0.95} & \textbf{31.35} & \textbf{12.84} & \textbf{1.25} & 7.79 & {6.31} & \textbf{0.92}  \\
  \bottomrule
  \end{tabular}}
  \caption{Quantitative results on the LRS2 test set and HDTF dataset. Please see \cref{app:additional_results} for more results.}
  \label{tab:results}
\end{table*}

\subsection{Lip Adaptation Model}

At this stage, we aim to adapt the lips to the given audio to achieve synchronized lips.
The generator $G_L$ takes the output from $G_E$, which is a face image with neutral mouth region.
We encode this image using an image encoder composed of several consecutive convolutional layers, batch normalization~\cite{ioffe2015batch}, and ReLU activation functions~\cite{nair2010rectified,krizhevsky2012imagenet}: $f_I^{512 \times 16 \times 16} = E_I(I)$.
Additionally, $G_L$ receives the corresponding audio snippet as a condition for adapting the lips.
We encode the mel-spectrogram representation of the audio using an audio encoder, that has similar architecture with ~\cite{prajwal2020lip}, into $f^{1 \times 1 \times 512}$ feature vector and incorporate it into the network via Adaptive Instance Normalization (AdaIN) layer~\cite{huang2017arbitrary}, which has shown more efficient performance~\cite{cheng2022videoretalking,zhong2023identity}.
However, we empirically find that using only AdaIN to feed audio into the network results in suboptimal lip-sync performance. 
To address this, we inject the audio into the embedding space as well by concatenating the encoded image and audio features along the depth dimension.
Our generator involves SPADE layers that help preserving identity better by retaining identity-specific details since we provide the original image as semantic input.
Moreover, we employ Pixel shuffling layers~\cite{shi2016real}, as it demonstrates better generation quality and tends to cause less artifacts compared to transposed convolution layers (see \cref{app:lip_adaptation_model} for architectural details).

To train our model, we use adversarial loss ($L_{GAN}$), perceptual loss ($L_{per}$), lip synchronization loss ($L_{ads}$) (see \Cref{sec:lip_synchronization}), and L1 pixel reconstruction loss ($L_{pixel}$):
\begin{equation}
    L = L_{GAN} + \lambda_1 L_{per} + \lambda_2 L_{ads} + \lambda_3 L_{pixel}
\end{equation}
where we empirically choose coefficients as follows: $(\lambda_1, \lambda_2, \lambda_3) = (4, 0.5, 10)$.

\subsection{Audio-Lip Synchronization} \label{sec:lip_synchronization}
The most common approach for learning audio-lip synchronization is to utilize the pretrained SyncNet model~\cite{prajwal2020lip} for audio-visual feature extraction to calculate synchronization loss. 
Recent studies~\cite{muaz2023sidgan,wang2023seeing,yaman2024audio,yaman2024audiodriventalkingfacegeneration} highlight fundamental issues with this approach and propose alternative methods.
Following \cite{yaman2024audiodriventalkingfacegeneration}, we employ a modified version of stabilized synchronization loss, which we refer to as the adapted stabilized synchronization loss, during the training of our lip adaptation model.
In the stabilized synchronization loss, the difference in similarity between (GT lips, audio) and (generated lips, audio) is utilized instead of solely relying on the similarity of the (generated lips, audio) pair.
Additionally, the similarity of the (reference lips, audio) pair is employed to adjust the loss when the reference shows a higher similarity to the audio.
However, since we don't use any reference image in our approach, we by-pass the similarity of the (reference lips, audio) pair and use the difference in similarity between (GT lips, audio) and (generated lips, audio) as follows:
\begin{equation}
    L_{ads} = -log(1 - |D(F^A, F^{I'}) - D(F^A, F^{I^{GT}})|)
\end{equation}
where $D$ is cosine similarity, $F^A$ indicates audio features, $F^{I'}$ and $F^{I^{GT}}$ represent generated image and ground-truth image features, respectively.
We obtain these features from SyncNet~\cite{prajwal2020lip} audio and image encoders.

\subsection{Training Strategy}
\label{sec:method:training}
First, we train our landmark generation model using a subset of the LRS2 dataset~\cite{LRS2}, selecting faces with closed lips by computing the distance between top and bottom lip landmarks. 
This step is crucial, as accurate lip landmark prediction is essential for guiding the face editing model ($G_E$) to have neutral lips.
Since this is a relatively straightforward approach, the subset of the LRS2 dataset is sufficient for learning generalized landmark prediction model for neutral lips.
In the second step, we use our landmark generator to produce neutral lip landmarks for each face in the LRS2 dataset. 
Then, we train the face editing model (see \Cref{sec:face_editing_model}) by conditioning it on the input image ---a face from the LRS2 dataset--- and the predicted neutral lip landmark map.
Finally, after rendering a face image with neutral mouth, we use it as an input image, along with the corresponding audio, in our lip adaptation model to generate final output, which is the face images with accurate lip movements regarding to the input audio.

While we process one face image per step ($T=1$) in the landmark generation and face editing models, we use $5$ images per step ($T=5$) in the our lip adaptation model, as maintaining temporal sequence is essential for achieving accurate lip synchronization as well as measuring it more efficiently during the training.
We use FAN~\cite{bulat2017far} to detect faces and apply tight cropping, adding $10\%$ margin at the bottom since FAN tends to cut off a small portion of the chin.
Given the low resolution of faces in LRS2 dataset, our model takes input images $128 \times 128$ resolution image as input.
Our audio encoder requires a mel-spectrogram of size $16 \times 80$, which derived from $16$ $kHz$ audio with a window size of $800$ and a hop size of $200$. 
We employ the Adam optimizer with $( \beta_1, \beta_2) = (0.5, 0.999) $. 
We set the learning rate to $1 \times 10^{-4}$ for all models.
We train our models on a single NVIDIA RTX A6000 GPU.

\noindent\textbf{Inference.}
During inference, our landmark predictor takes only the landmarks of the input image and generates a landmark map with a neutral mouth.
Then, the face editing model utilizes the input image and the predicted landmark map to modify the mouth region accordingly.
Finally, the lip adaptation model processes the output of the face editing model along with the input audio to adjust the lip movements.
In summary, our entire pipeline requires only a single image during inference.
Some might argue that certain existing models (e.g., Wav2Lip~\cite{prajwal2020lip}, VRT~\cite{cheng2022videoretalking}) can also rely solely on the input image during inference by selecting the input image and identity reference as the same.
While this is technically possible, it applies only during inference, not during training.
Consequently, identity reference-related issues during training persist.
Moreover, these models still require a masked input image in both training and inference, leading to all the previously identified mask-related problems.
Last but not least, empirical results indicate that the identity reference influences the lip-sync performance of these models~\cite{cheng2022videoretalking,muaz2023sidgan,yaman2024audiodriventalkingfacegeneration} (See Section \ref{sec:intro} and performance degradation from Table \ref{tab:results} to Table \ref{tab:cross_test}).

\begin{figure*}[tb]
  \centering
  \begin{subfigure}{0.105\textwidth}
    \includegraphics[trim={0cm 2cm 0cm 2cm},clip,width=\textwidth]{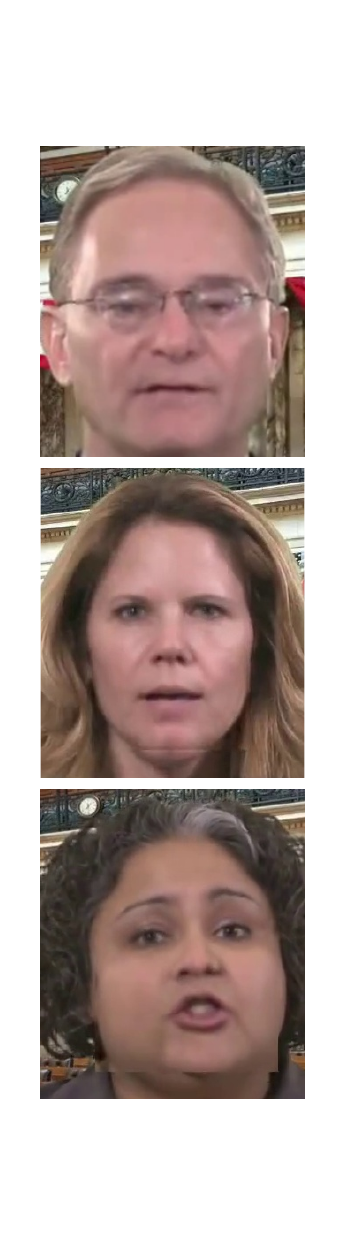}
    \caption{Wav2Lip}
    \label{fig:wav2lip}
  \end{subfigure}
  \hspace{-0.5cm}
  \begin{subfigure}{0.105\textwidth}
    \includegraphics[trim={0cm 2cm 0cm 2cm},clip,width=\textwidth]{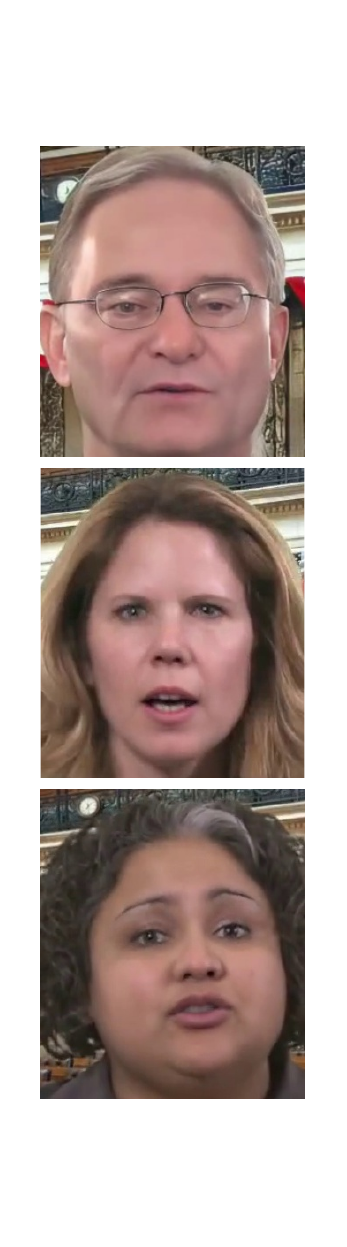}
    \caption{VRT}
    \label{fig:VRT}
  \end{subfigure}
  \hspace{-0.5cm}
  \begin{subfigure}{0.105\textwidth}
    \includegraphics[trim={0cm 2cm 0cm 2cm},clip,width=\textwidth]{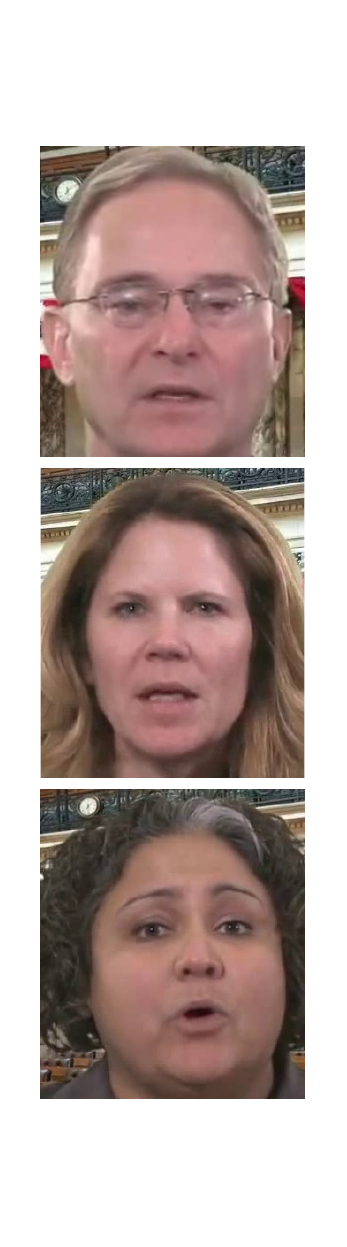}
    \caption{DINet}
    \label{fig:DINet}
  \end{subfigure}
  \hspace{-0.5cm}
  \begin{subfigure}{0.105\textwidth}
    \includegraphics[trim={0cm 2cm 0cm 2cm},clip,width=\textwidth]{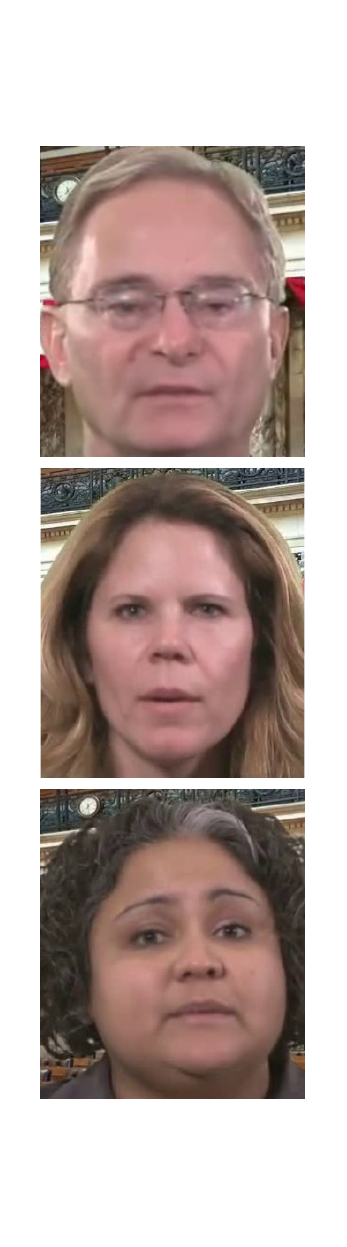}
    \caption{IPLAP}
    \label{fig:IPLAP}
  \end{subfigure}
  \hspace{-0.5cm}
  \begin{subfigure}{0.105\textwidth}
    \includegraphics[trim={0cm 2cm 0cm 2cm},clip,width=\textwidth]{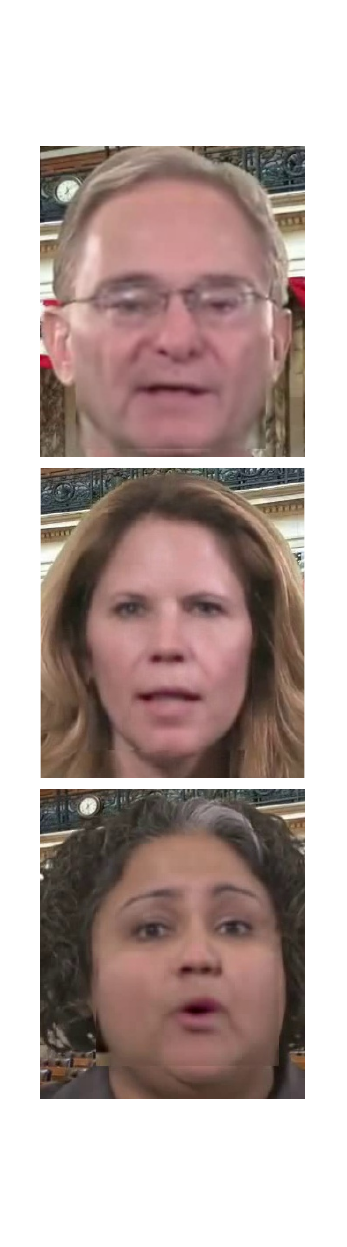}
    \caption{TalkLip}
    \label{fig:TalkLip}
  \end{subfigure}
  \hspace{-0.5cm}
  \begin{subfigure}{0.105\textwidth}
    \includegraphics[trim={0cm 2cm 0cm 2cm},clip,width=\textwidth]{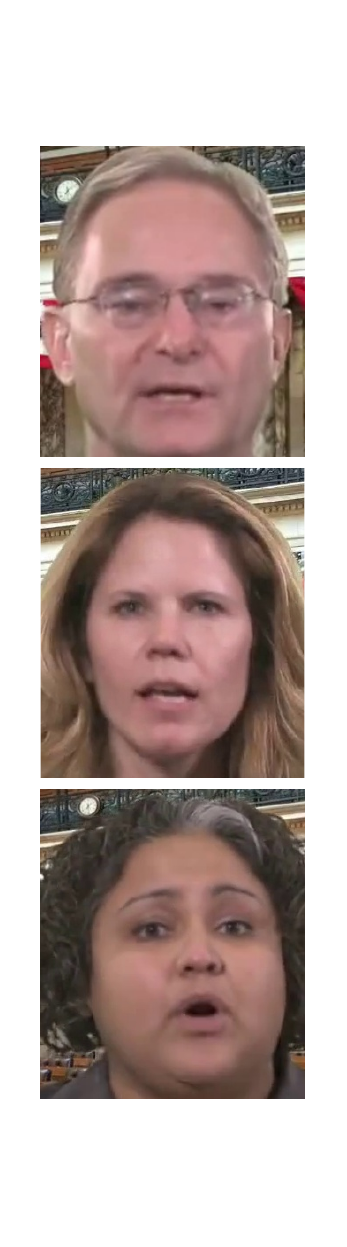}
    \caption{AVTFG}
    \label{fig:AVTFG}
  \end{subfigure}
  \hspace{-0.5cm}
  \begin{subfigure}{0.105\textwidth}
    \includegraphics[trim={0cm 2cm 0cm 2cm},clip,width=\textwidth]{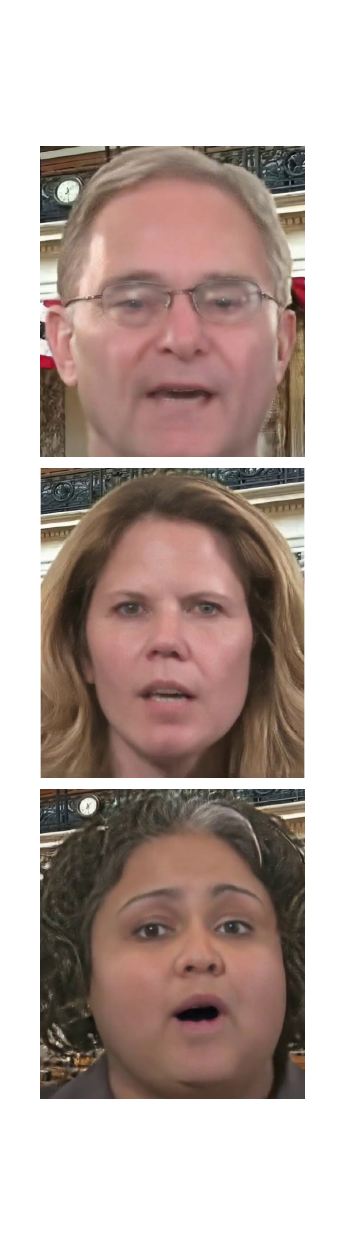}
    \caption{PLGAN}
    \label{fig:PLGAN}
  \end{subfigure}
  \hspace{-0.5cm}
  \begin{subfigure}{0.105\textwidth}
    \includegraphics[trim={0cm 2cm 0cm 2cm},clip,width=\textwidth]{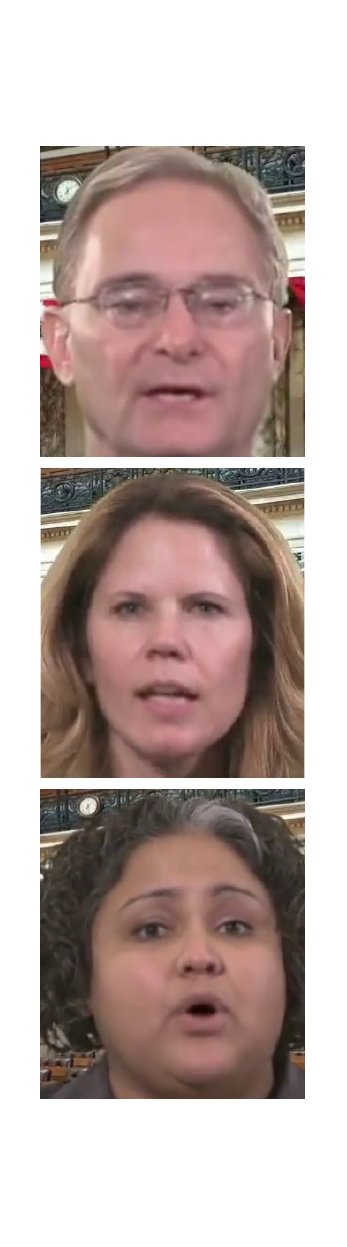}
    \caption{Diff2Lip}
    \label{fig:diff2lip}
  \end{subfigure}
  \hspace{-0.5cm}
  \begin{subfigure}{0.105\textwidth}
    \includegraphics[trim={0cm 2cm 0cm 2cm},clip,width=\textwidth]{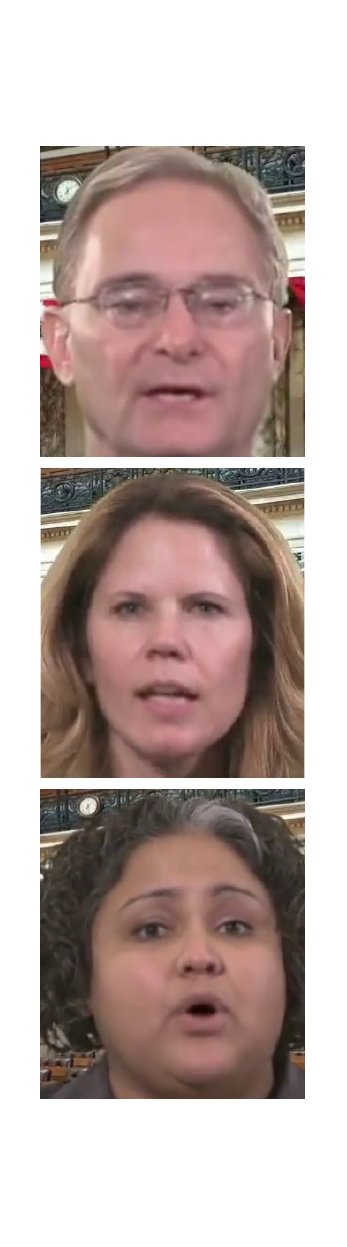}
    \caption{Ours}
    \label{fig:Ours}
  \end{subfigure}
  \hspace{-0.5cm}
  \begin{subfigure}{0.105\textwidth}
    \includegraphics[trim={0cm 2cm 0cm 2cm},clip,width=\textwidth]{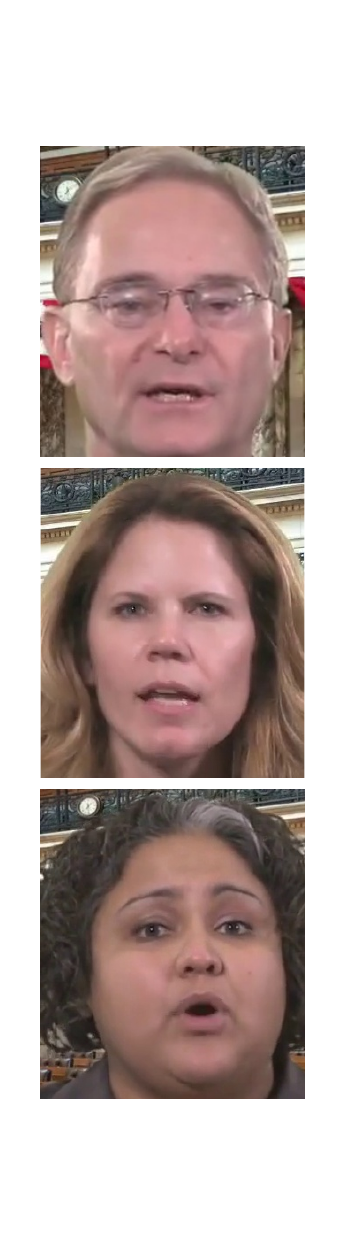}
    \caption{GT}
    \label{fig:GT}
  \end{subfigure}
  
  \caption{Qualitative comparison of our model with SOTA methods. The samples are randomly selected from generated videos in the unseen HDTF dataset. For more qualitative comparison, please check \cref{app:additional_results} and Supplementary videos.}
  \label{fig:qualitative_comparison}
\end{figure*}

\begin{table}[t]
    \centering
    \resizebox{\linewidth}{!}{\begin{tabular}{l|cccccc}
    Method & SSIM & PSNR & FID & LSE-C & LSE-D & CSIM \\
    \midrule
    Wav2Lip & 0.842 & 25.835 & 7.89 & {7.347} & 7.184 & 0.736 \\
    VideoReTalking & 0.837 & 26.539 & 9.75 & 6.815 & 7.743 & 0.749 \\
    DINet & 0.776 & 24.034 & 4.17 & 4.461 & 9.554 & 0.724 \\
    TalkLip & 0.849 & 25.701 & 4.04 & 6.044 & 8.206 & 0.739 \\
    IPLAP & 0.861 & {28.989} & 3.95 & 3.627 & 10.102 & 0.766 \\
    AVTFG & 0.849 & 26.425 & 5.78 & 6.844 & 7.901 & 0.723 \\
    PLGAN & 0.855 & 25.376 & 4.11 & \textbf{7.578} & \textbf{6.805} & 0.731 \\
    Diff2Lip & 0.916 & 30.317 & 3.59 & 6.710 & 7.261 & 0.833 \\
    \midrule
    Ours & \textbf{0.924} & \textbf{31.472} & \textbf{3.52} & 6.525 & 7.388 & \textbf{0.842} \\
    \bottomrule
    \end{tabular}}
    \caption{Quantitative results on the LRS2 test set for cross matching scenario (random video--audio pairs).}
    \label{tab:cross_test}
\end{table}

\section{Experimental Results}

\noindent\textbf{Datasets}
We trained our landmark prediction model on a subset of the LRS2 dataset~\cite{LRS2} and the other two models on the entire LRS2 dataset. 
We evaluated our overall approach using both the LRS2 test set and the HDTF dataset~\cite{zhang2021flow}.

\noindent\textbf{Baseline and Evaluation}
We select state-of-the-art methods in 2D audio-driven talking face generation for comparison with our model and follow established evaluation metrics from the literature~\cite{prajwal2020lip,cheng2022videoretalking,zhang2023dinet,zhong2023identity,wang2023seeing,yaman2024audiodriventalkingfacegeneration,mukhopadhyay2024diff2lip}.
For visual quality assessment, we employed SSIM~\cite{wang2004image}, PSNR, and FID~\cite{heusel2017gans}, while for lip-sync evaluation, we used LSE-C, LSE-D~\cite{chung2017out,prajwal2020lip}, and LMD~\cite{chen2019hierarchical}.
To evaluate how well the models preserve identity, we employed the CSIM, %metric
which measures cosine similarity between the features of the generated and target faces.
For feature extraction, we used the pretrained ArcFace model~\cite{deng2019arcface} (see \cref{app:evaluation_metrics} for details).
We share various ablation studies in \cref{sec:ablation_study} and \cref{app:ablation_study}.

\subsection{Quantitative Results}

In \Cref{tab:results}, we present quantitative results on the LRS2 test set and the HDTF dataset.
This is using the standard approach for evaluating talking face generation, i.e., videos are generated with their respective GT audio, allowing us to measure the performance accurately even for metrics that require exact GT data.
We outperform other methods in visual quality metrics across both datasets, and our mask-free approach enables significantly better identity preservation, as reflected in the CSIM scores.
In terms of lip-sync accuracy, our model demonstrates comparable performance. %in %both visual quality and 
We consistently outperform other approaches in our user study (see \cref{app:user_study}).
In contrast to this, \Cref{tab:cross_test} demonstrates quantitative results on the LRS2 test set for cross audio-video pairs, i.e., randomly pairing videos and audio in the test set. 
This is done to eliminate any potential lip leakage, following the setup in Wav2Lip~\cite{prajwal2020lip}.
Note that the LSE-C \& -D metrics do not require any GT data, whereas the remaining metrics do. 
Although these models alter the mouth region, we still use the input images to measure these metrics, as the models are expected to preserve various details regardless.
The results clearly show the effectiveness of our approach, especially in visual quality and identity preservation.
We achieve the best SSIM, PSNR and FID scores and competitive performance on LSE-C, and LSE-D. 
As in \Cref{tab:results}, we again reach the best CSIM, highlighting the strong identity preservation capability of our method.
The performance of the other methods mostly deteriorates notably.

\subsection{Qualitative Results}
We use generated videos from the HDTF dataset to qualitatively evaluate the performance of our method alongside other approaches.
In \Cref{fig:qualitative_comparison}, we present results from recent SOTA models. % with publicly available checkpoints.
Our model generates lip shapes that align most accurately with the GT data.
Although Wav2Lip, TalkLip, AVTFG, and PLGAN demonstrate comparable performance, they do not achieve the same level of accuracy as our model.
Moreover, DINet's outputs closely resemble the GT lip shapes, however, it was trained on HDTF dataset, unlike the other methods.
Additionally, TalkLip and Wav2Lip occasionally exhibit artifacts along the facial borders, especially near the lower edge, while DINet and VideoReTalking do not perform as well as our model in preserving identity.
Similarly, despite its high quality, Diff2Lip exhibits noticeable teeth artifacts.
Moreover, PLGAN shows artifacts in teeth generation.

\subsection{Ablation Study}
\label{sec:ablation_study}

\subsubsection{Analysis of Landmark Prediction}

\begin{figure}[t]
  \centering
  \begin{subfigure}{\linewidth}
    \includegraphics[width=\linewidth]{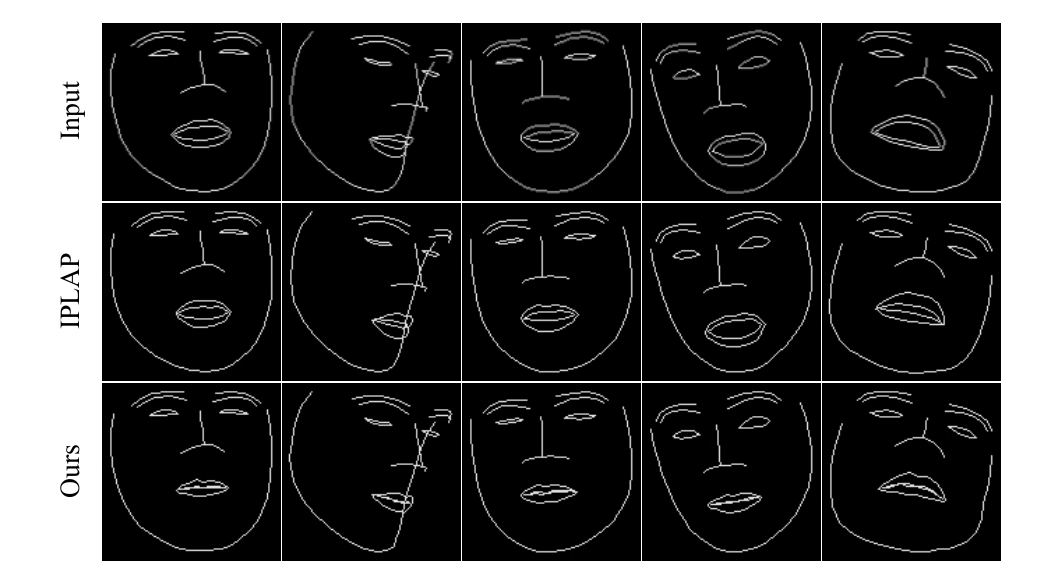}
    \vspace{-0.65cm}
    \caption{Input maps and predicted neutral-mouth landmark maps}
    \label{fig:comparison_of_landmark_and_rendering:generated_landmarks}
  \end{subfigure}
  
  \begin{subfigure}{\linewidth}
    \includegraphics[trim={0cm 0cm 0cm 0cm},clip,width=\linewidth]{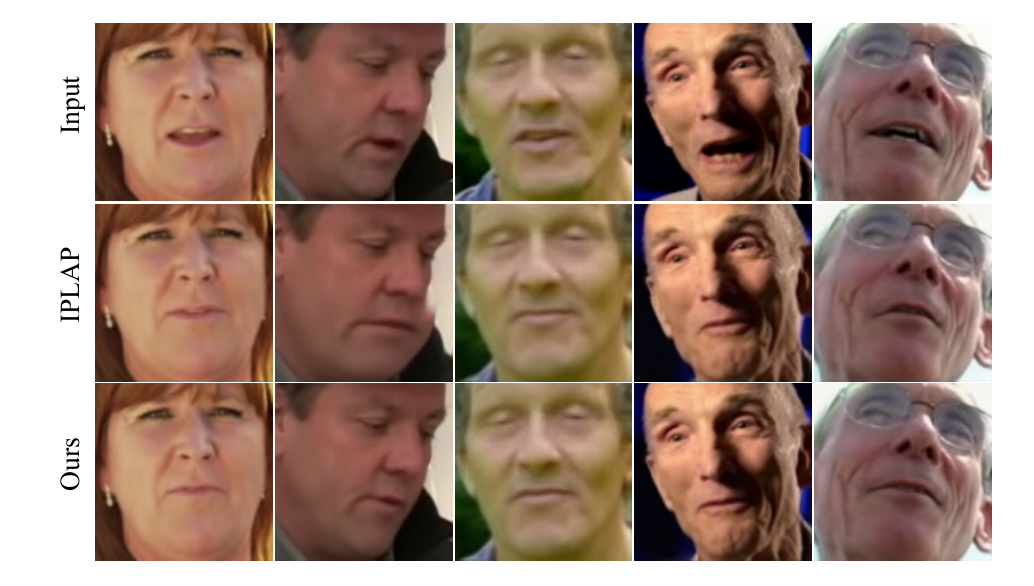}
    \vspace{-0.65cm}
    \caption{Input images and generated faces with neutral mouth.}
    \label{fig:comparison_of_landmark_and_rendering:rendered_images}
  \end{subfigure}
  
  \caption{Comparison of the IPLAP landmark generation method (with silent audio) and our landmark predictor. 
  }
  \label{fig:comparison_of_landmark_and_rendering}
\end{figure}

In \Cref{tab:landmark_comparison}, we conduct experiments to evaluate our $T_L$ model's performance in generating landmarks for a neutral mouth.
For comparison, we use IPLAP landmark prediction model with silent audio, expecting it to generate a neutral mouth since no speech is present.
We assess the models' performance with three metrics.
$LD_{full}$ calculates the L2 distance between the generated and ground-truth landmarks, while $LD_{lip}$ measures the L2 distance specifically between the generated and ground-truth lip landmarks.
The final metric, $LD_{c}$, represents the vertical L1 distance between the center point of the upper and lower lips, which is expected to be minimal in a neutral mouth scenario.
In all three metrics on two different datasets, we clearly surpass IPLAP landmark generator for generating more accurate neutral mouth when there is no speech. % We also validate this qualitatively.
Ours w/o lip loss also validates the usefulness of our dedicated lip landmark loss in the training.
Note, however, that our $T_L$ is specifically trained for generating neutral mouths, in contrast to IPLAP.
For qualitative comparison, \Cref{fig:comparison_of_landmark_and_rendering} presents the predicted landmark maps generated by our landmark predictor alongside those from the IPLAP predictor when using silent audio.
Our model demonstrates superior performance in achieving a neutral-mouth position (see \cref{fig:comparison_of_landmark_and_rendering:generated_landmarks}).
Additionally, our model produces images with a more accurate closed-mouth appearance (see \cref{fig:comparison_of_landmark_and_rendering:rendered_images}).
Overall, our approach achieves higher accuracy in neutralizing the mouth while preserving the given pose.

\begin{table}[t]
    \centering
    \resizebox{\linewidth}{!}{\begin{tabular}{l|ccc|ccc}
    & \multicolumn{3}{c}{LRS2} & \multicolumn{3}{c}{HDTF} \\
    \midrule
    Method &  LD$_{full}$ $\downarrow$ & LD$_{lip}$ $\downarrow$ & LD$_{c}$ $\downarrow$ & LD$_{full}$ $\downarrow$ & LD$_{lip}$ $\downarrow$ & LD$_{c}$ $\downarrow$ \\
    \midrule
    IPLAP w/ silent audio & 9.504 & 2.501 & 0.305 & 10.388 & 2.755 & 0.322 \\
    \midrule
    Ours w/o lip loss & 9.694 & 2.580 & 0.329 & 10.541 & 2.806 & 0.340 \\
    Ours & \textbf{9.418} & \textbf{2.459} & \textbf{0.293} & \textbf{10.199} & \textbf{2.657} & \textbf{0.301} \\
    \bottomrule
    \end{tabular}}
    \caption{Quantitative results of our landmark predictor and IPLAP landmark predictor on the LRS2 test set and HDTF dataset.}
    \label{tab:landmark_comparison}
\end{table}

\subsubsection{Face Generation with Neutral Mouth}

\begin{figure}[t]
    \centering
    \includegraphics[width=1\linewidth]{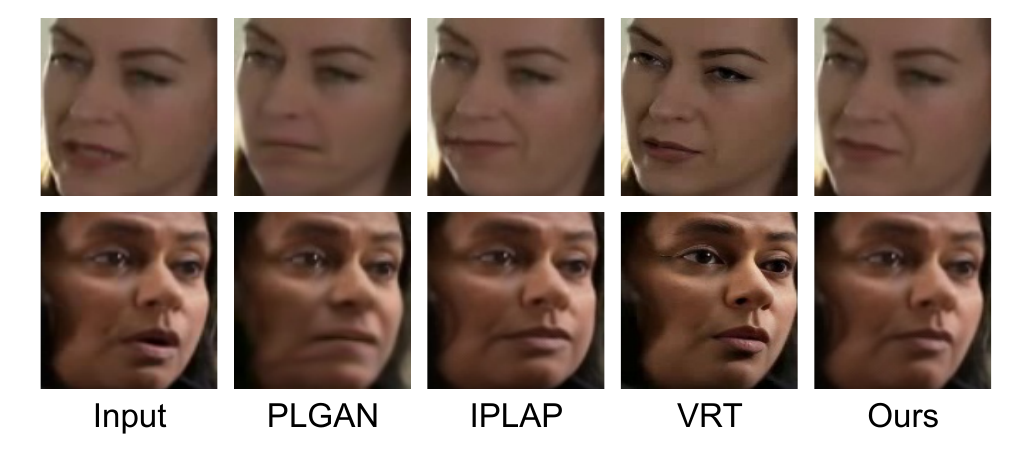}
    \vspace{-0.8cm}
    \caption{Generated samples with neutral mouth by different methods. The samples are from LRS2 test set.}
    \label{fig:ablation_study_silent}
\end{figure}

In \Cref{tab:silent_models}, we generate faces with neutral mouths on the LRS2 test set using our $G_E$ face editing model.
We also test the canonical face generation model of VideoReTalking, which generates faces by neutralizing both expression and mouth position, and silent-lip generator from PLGAN.
For evaluation, we employ visual quality and identity preservation metrics.
According to the scores, our model clearly surpasses VideoReTalking and IPLAP models in both visual quality and identity preservation metrics.
Despite comparable performance of PLGAN on PSNR and FID, we outperform it in SSIM and CSIM (see \Cref{fig:ablation_study_silent}).

In \Cref{tab:results_w_different_silent_models}, we use aforementioned neutral mouth generation method in place of our first and second stages.
We then train our lip adaptation model with these images to explore the impact of different neutral mouth generation models on talking face generation.
Our model achieves the best performance across all metrics.
The highest CSIM scores clearly demonstrate that $G_E$ preserves identity while generating face image with a neutral / closed mouth.

\subsubsection{Masking Strategy}

In \Cref{tab:masking_ablation}, we compare our approach with a masking-based baseline approach, where we incorporate a masking strategy in the lip adaptation model ($G_L$) and omit the first and second stages.
Due to the masking, we utilize a randomly selected identity reference image.
In the second experiment, we apply our full setup but mask the input image in the second stage. 
Therefore, we again provide identity reference.
The output image from the second stage, a face with a neutral mouth, is then used as the identity reference in the third stage, where we also mask the input image.
As expected, this second approach outperforms the first (baseline), as the neutral identity reference strategy has already been validated in PLGAN.
However, our mask-free approach clearly demonstrates the best performance across all metrics.

\begin{table}[t]
    \footnotesize
    \centering
    \begin{tabular}{l|cccc}
    Method & SSIM & PSNR & FID & CSIM \\
    \midrule
    VideoReTalking & 0.646 & 22.12 & 33.60 & 0.603 \\
    IPLAP w/ silent audio & 0.859 & 28.45 & 6.78 & 0.821  \\
    PLGAN & 0.908 & \textbf{30.32} & \textbf{4.41} & 0.856 \\ 
    \midrule
    Ours & \textbf{0.912} & 29.74 & 4.92 & \textbf{0.887} \\
    \bottomrule
    \end{tabular}
    \caption{Quantitative comparison of our face editing model for neutral mouth generation with the canonical face generation model from VideoReTalking, the IPLAP model with silent audio, and PLGAN silent-lip generation model. }
    \label{tab:silent_models}
\end{table}

\begin{table}[t]
    \centering
    \setlength{\tabcolsep}{3pt}
    \resizebox{\linewidth}{!}{\begin{tabular}{l|ccccccc}
    Method &  SSIM  & PSNR & FID  & LMD  & LSE-C & LSE-D  & CSIM  \\
    \midrule
    w/ VideoRetalking & 0.77 & 24.08 & 5.45 & 2.50 & 7.03 & 6.88 & 0.76 \\
    w/ IPLAP          & 0.83 & 28.85 & 3.77 & 1.92 & 7.72 & 6.32 & 0.82 \\
    w/ PLGAN          & 0.79 & 23.63 & 4.32 & 2.58 & 7.13 & 6.83 & 0.74 \\
    w/ our model      & \textbf{0.95} & \textbf{33.96} & \textbf{3.57} & \textbf{1.18} & \textbf{7.76} & \textbf{6.32} & \textbf{0.88} \\
    \bottomrule
    \end{tabular}}
    \caption{Ablation study of neutral mouth generation methods. We use different neutral mouth generation models to synthesize faces with neutral mouth and train our lip adaptation model with them to explore their effects on the final performance.}
    \label{tab:results_w_different_silent_models}
\end{table}

\begin{table}[t]
    \centering
    \setlength{\tabcolsep}{3pt}
    \resizebox{\linewidth}{!}{\begin{tabular}{l|cccccccc}
    Method &  SSIM & PSNR  & FID  & LMD  & LSE-C  & LSE-D  & CSIM & Ep. \\
    \midrule
    Baseline & 0.81 & 25.28 & 14.89 & 2.41 &  7.61 & 6.45 & 0.75 & 120 \\
    Ours w/ masking  & 0.85 & 27.41 & 7.94 & 2.04 & 7.79 & 6.31 & 0.76 & 58 \\
    Ours (Mask-Free)  & \textbf{0.95} & \textbf{33.96} & \textbf{3.57} & \textbf{1.18} & \textbf{7.76} & \textbf{6.32} & \textbf{0.88} & 32 \\
    \bottomrule
    \end{tabular}}
    \caption{Ablation study for masking approach.}
    \label{tab:masking_ablation}
\end{table}

\section{Conclusion}
We introduce a mask-free approach for talking face generation.
First, we transform the input video frames to have neutral, closed lips using a two-stage landmark-based face editing model trained with unpaired data.
Then, we apply an audio-conditioned lip adaptation model on the transformed sequence of neutral-mouth faces to generate lips matching the given audio.
Our experiments show that MF-Talk achieves competitive results on LRS2 and HDTF, especially preserving identity better than masking-based approaches, and the extensive ablation studies underline the importance of each pipeline component.

\noindent\textbf{Limitations \& Ethics.} 
Our model generates suboptimal teeth due to having neutral/closed mouth in the input of the lip adaptation model (e.g., no visible teeth in the input).
This occasionally conceals the subject's teeth. 
However, relevant information may still exist in the feature space, allowing the model to accurately generate the teeth according to our empirical observation.
Generating lip-sync faces offers valuable applications but is vulnerable to misuse, such as in deepfake creation. We will implement watermarking to prevent unauthorized use of our model.

{
    \small
    \bibliographystyle{ieeenat_fullname}
    \bibliography{main}
}

\clearpage
\appendix
\clearpage
\setcounter{page}{1}

\maketitlesupplementary

\section{Datasets}
\label{app:dataset}

\textbf{LRS2.} This dataset comprises $45839$ utterances in the training set, with $1082$ and $1243$ utterances in the validation and test sets, respectively.
Each utterance is a short clip approximately $2$ seconds long.

\noindent\textbf{HDTF.} This dataset consists of $174$ relatively long, high-quality video clips that feature various subjects.

\section{Method}
\label{app:method}

\subsection{Landmark Prediction Model}
\label{app:landmark_prediction_model}

\paragraph{How do we prepare training data?}
In order to train our face model, we need ground-truth images with a closed or neutral mouth. 
Therefore, we select faces from the LRS2 training set that have a closed mouth. 
This selection is based on the calculation of the distance between the landmark points of the upper and lower lips. 
Using this data, we then train our landmark prediction model ($T_L$).

\paragraph{Training setup}
In our model, we represent facial landmark points as a 1D vector. 
We use $k$ previous face frames, detect their landmark points, and encode them in vector format. 
These previous frames help capture identity-related details at the landmark level. 
Additionally, we provide the upper-face landmarks from the current time step $t$. 
However, we do not include lower-face landmarks, as our model is designed to learn and predict them, ensuring they represent a neutral or closed-mouth expression.
In the selected subset, we have a diverse range of poses, including some very challenging ones. 
Moreover, the task is relatively easier since it involves only predicting the lower-face landmark points representing a neutral mouth. 
These predictions must also maintain coherence with the upper-face landmarks and the person's identity (e.g., mouth and cheek size), which is derived from previous frames.

\paragraph{CNN Encoder}
In our network, each CNN encoder has 20 consecutive 1D convolutional layers, producing $1 \times 512$ embeddings.

\paragraph{Landmark distance loss}
The utilized landmark distance loss ensures that the model accurately reconstructs the upper-face landmarks (which are already provided as input) and predicts the lower-face landmarks. 
This includes both correctly modeling neutral or closed-mouth landmarks and properly localizing them by maintaining coherence with the upper-face landmarks and the overall pose of the face.

\paragraph{How can we use this model in inference?}
During inference, we similarly provide the previous $k$ frames and the upper-face landmarks of the current frame to predict the full set of landmarks. 
No neutral face landmarks are required as input during inference, as our model can generate neutral face landmarks from any given input.

\paragraph{Performance}
The experimental results on the LRS2 test data clearly demonstrate that our landmark generator accurately predicts neutral mouth landmarks while maintaining coherence with the rest of the face.

\subsection{Landmark-driven Face Editing Model}
\label{app:landmark_driven_face_editing_model}

\paragraph{Training setup}
In this model, we take a face input along with a landmark map drawn from the predicted landmark vector. 
This vector is generated by our landmark prediction model ($T_L$) based on the original landmarks of the input image. 
While the predicted landmarks closely resemble the original ones, the lower-face landmarks are modified to represent a neutral mouth. 
Our face editing model ($G_E$) is responsible for applying these lower-face modifications at the RGB image level, conditioned on the input landmark map.

\paragraph{Performance Analysis}
Since our face editing model ($G_E$) does not use a masking strategy, it avoids the mask-related issues mentioned earlier.
Another important aspect is identity preservation. 
The experimental results clearly demonstrate that our face editing model ($G_E$) preserves identity with high accuracy. 
In addition, the visual quality remains very accurate. 
By utilizing feature matching loss and perceptual loss, and with the absence of masking (which eliminates information loss), the task becomes largely about reconstructing the input with slight modifications. 
As a result, our model can both accurately preserve identity and deliver high visual quality performance.

\paragraph{Architecture}
We present our architectural design for face encoder and face decoder in Table \ref{tab:face_encoder} and Table \ref{tab:face_decoder_face_editing}, respectively.

\subsection{Mouth Classification Model}
\label{app:mouth_classification_model}
We finetune pretrained ResNet-50 model (which was trained on ImageNet dataset) on LRS2 dataset with Binary Cross-Entropy Loss.
We label data open and closed mouths as in $T_L$ training.
The model achieved 89.06\% classification accuracy on LRS2 test set.

\subsection{Lip Adaptation Model}
\label{app:lip_adaptation_model}

\paragraph{Architecture}
We present the face encoder architecture in Table \ref{tab:face_encoder}. 
After each convolutional layer, we utilize batch normalization and ReLU activation function.
Please note that we choose the same architecture design for the face encoder in the face editing model ($G_E$) and the lip adaptation model ($G_L$).
We introduce the details of the face decoder in Table \ref{tab:face_decoder_lip_adaptation}.

\begin{table}[b]
    \centering%\scriptsize
    %\vspace{-1cm}
    \setlength{\tabcolsep}{5.2pt}
    %\resizebox{\linewidth}{!}{
    \begin{tabular}{c|c|c}
    Layer Name & Output Size & Layer Detail \\
    \midrule
    Conv$_1$ & $ 128 \times 128 \times 64 $ & $ [7 \times 7, 64] $, stride 1 \\
    \midrule
    Conv$_2$ & $ 64 \times 64 \times 128 $ & $ [3 \times 3, 128] $, stride 2 \\
    \midrule
    Conv$_3$ & $ 32 \times 32 \times 256 $ & $ [3 \times 3, 256] $, stride 2 \\
    \midrule
    Conv$_4$ & $ 16 \times 16 \times 512 $ & $ [3 \times 3, 512] $, stride 2 \\
    \bottomrule
    \end{tabular}
    \caption{Architecture of the face encoder in the face editing model ($G_E$) and the lip adaptation model ($G_L$). After each convolutional layer, we employ batch normalization (BN) and ReLU activation function.}
    \label{tab:face_encoder}
\end{table}

\begin{table}[b]
    \centering%\scriptsize
    %\vspace{-1cm}
    \setlength{\tabcolsep}{5.2pt}
    \resizebox{\linewidth}{!}{
    \begin{tabular}{c|c|c}
    Layer Name & Output Size & Layer Detail \\
    \midrule
    SPADE$_1$ & $16 \times 16 \times 512$  & channel 512, modulation channel 3\\
    \midrule
    PixelShuffle$_1$ & $32 \times 32 \times 128$ & upscale factor 2 \\
    \midrule
    SPADE$_2$ & $32 \times 32 \times 128$ & channel 128, modulation channel 3 \\
    \midrule
    PixelShuffle$_2$ & $64 \times 64 \times 32$ & upscale factor 2 \\
    \midrule
    SPADE$_3$ & $64 \times 64 \times 32$ & channel 32, modulation channel 3 \\
    \midrule
    PixelShuffle$_3$ & $128 \times 128 \times 8$ & upscale factor 2 \\
    \midrule
    Conv$_4$ & $128 \times 128 \times 3$ & $ [7 \times 7, 3], stride 1 $ \\
    \bottomrule
    \end{tabular}}
    \caption{Architecture of the face decoder in the face editing model ($G_E$).}
    \label{tab:face_decoder_face_editing}
\end{table}

\begin{table}[b]
    \centering%\scriptsize
    %\vspace{-1cm}
    \setlength{\tabcolsep}{5.2pt}
    \resizebox{\linewidth}{!}{
    \begin{tabular}{c|c|c}
    Layer Name & Output Size & Layer Detail \\
    \midrule
    Conv$_1$ & $16 \times 16 \times 512$ & $[3 \times 3, 512]$, stride 1 \\
    \midrule
    SPADE$_1$ & $16 \times 16 \times 512$  & channel 512, modulation channel 3\\
    \midrule
    AdaIN$_1$ & $16 \times 16 \times 512$ & input channel 512, modulation channel 512 \\
    \midrule
    PixelShuffle$_1$ & $32 \times 32 \times 128$ & upscale factor 2 \\
    \midrule
    SPADE$_2$ & $32 \times32 \times 128$ & channel 128, modulation channel 3 \\
    \midrule
    AdaIN$_2$ & $32 \times 32 \times 128$ & input channel 128, modulation channel 512 \\
    \midrule
    PixelShuffle$_2$ & $64 \times 64 \times 32$ & upscale factor 2 \\
    \midrule
    SPADE$_3$ & $64 \times 64 \times 32$ & channel 32, modulation channel 3 \\
    \midrule
    PixelShuffle$_3$ & $128 \times 128 \times 8$ & upscale factor 2 \\
    \midrule
    Conv$_4$ & $128 \times 128 \times 3$ & $ [7 \times 7, 3], stride 1 $ \\
    \bottomrule
    \end{tabular}}
    \caption{Architecture of the face decoder in the lip adaptation model ($G_L$).}
    \label{tab:face_decoder_lip_adaptation}
\end{table}

\section{Evaluation Metrics}
\label{app:evaluation_metrics}

\noindent\textbf{Structural Similarity Index Measure (SSIM).}
This metric is for measuring the perceived quality.
We need to have ground truth images for this metric.
Higher score means more quality.

\begin{equation}
    SSIM(x,y) = \frac{(2 \mu_x \mu_y + c_1)(2 \sigma_{xy} + c_2)}{(\mu_x^2 + \mu_y^2 + c_1)(\sigma_x^2 + \sigma_y^2 + c_2)}
\end{equation}

\noindent\textbf{Peak Signal-to-Noise Ration (PSNR).} PSNR assesses visual quality by using the ratio of the maximum possible squared pixel value to the mean squared error (MSE) between the generated image and the ground truth. Higher values indicate better visual quality.

\begin{equation}
    PSNR(I', I) = 10 \ast log_{10} \frac{max(I')^2}{MSE(I', I)}
\end{equation}

\begin{equation}
    MSE(I', I) = \frac{1}{H W} \sum_{i=0}^{H-1} \sum_{j=0}^{W-1} |I'_{i,j} - I_{i,j}|^2
\end{equation}

\noindent\textbf{Fréchet Inception Distance (FID).} FID~\cite{heusel2017gans} measures visual quality by calculating the distance between generated images and ground truth images in feature space. 
A lower FID score, approaching zero, indicates better visual quality.
First, we extract features from real images and generated samples using the last pooling layer of the pre-trained Inception-V3 model~\cite{szegedy2016rethinking}, which has been trained on the large-scale ImageNet dataset~\cite{deng2009imagenet} for image classification.
FID formula is as follows:

\begin{equation}
    FID(F', F) = |\mu_{F'} - \mu_{F}| + TR(\Sigma_{F'} + \Sigma_{F} - 2 (\Sigma_{F'} \Sigma_{F})^{\frac{1}{2}})
\end{equation}

\noindent\textbf{Landmark Mouth Distance (LMD).} LMD is a metric for evaluating synchronization in videos using only visual data. 
Specifically, it involves detecting lip landmark points in both the generated samples and their ground truth counterparts~\cite{bulat2017far}\footnote{https://github.com/1adrianb/face-alignment}, then calculating the distance between them~\cite{chen2019hierarchical}. 
A smaller distance indicates greater similarity and better lip synchronization. 
However, LMD is not a robust metric for assessing synchronization, as variations in lip aperture and spreading can increase the distance despite maintaining synchronization.

\noindent\textbf{LSE-C and LSE-D.} LSE-C and LSE-D are metrics for evaluating synchronization between audio and lip movements in generated faces, measuring confidence and distance, respectively~\cite{chung2017out,prajwal2020lip}. 
SyncNet~\cite{chung2017out}, a network with jointly trained audio and image encoders, is used for extracting audio and visual features to assess synchronization. 
Higher LSE-C values and lower LSE-D values indicate better audio-visual synchronization.

\noindent\textbf{CSIM.}
This metric computes the cosine similarity between the generated face features and the ground truth (GT) face features. 
The features used are extracted from a pretrained ArcFace model~\cite{deng2019arcface}. 

\section{User Study and Runtime Analysis}
\label{app:user_study}

\begin{table}[b]
    \centering%\scriptsize
    %\vspace{-1cm}
    \setlength{\tabcolsep}{5.2pt}
    \resizebox{\linewidth}{!}{
    \begin{tabular}{l|cccc||cc}
    Method &  Sync  & Vis & Identity  & Overall  & Runtime* & Resolution \\
    \midrule
    Wav2Lip & 2.31 & 0.98 & 1.19 & 1.49 & 28.39 & 96$\times$96 \\
    DINet   & 1.47 & 1.95 & 1.84 & 1.75 & 129.85 & 128$\times$128 \\
    VRT     & 3.55 & 3.87 & 3.92 & 3.78 & 642.50 & 96$\times$96 \\
    TalkLip & 0.85 & 0.05 & 0.10 & 0.33 & 171.24 & 96$\times$96 \\
    IPLAP   & 2.71 & 3.71 & 3.96 & 3.46 & 420.46 & 128$\times$128 \\
    AVTFG   & 3.88 & 4.02 & 3.90 & 3.93 & 55.41  & 96$\times$96 \\
    PLGAN   & 4.12 & 3.79 & 3.95 & 3.96 & 371.59 & 96$\times$96 \\
    Ours    & \textbf{4.28} & \textbf{4.48} & \textbf{4.27} & \textbf{4.34} & 128.17 & 128$\times$128 \\
    \bottomrule
    \end{tabular}}
    \caption{User study for lip-sync, visual quality, and identity preservation and runing time analysis. Reported scores are MOS, scaled to $[0,5]$. * in sec / video min. Please consider the resolution.}
    \label{tab:user_study}
\end{table}

We conduct a user study to evaluate lip-sync accuracy, visual quality, and identity preservation.
Ten participants participated in the study and we randomly selected ten videos for each model from the HDTF dataset, which is unseen data for all models except DINet.
The results are presented in Table \ref{tab:user_study} and the scores indicate the mean opinion score (MOS), scaled to $ [0, 5] $.
We also analyze the running time of the models.
The results in Table \ref{tab:user_study} state that, despite the fact that it involves three submodules, our model achieves a relatively fast running time performance.

\section{Ablation Study}
\label{app:ablation_study}

\subsection{Masking Strategy}
We use three setups in the ablation study for masking. 
In the baseline setup, we redesign our lip adaptation model ($G_L$) in a traditional manner. 
It takes an identity reference, along with the audio and face inputs, and masks the lower half of the face input. 
This setup trains the model in the traditional way for talking face generation. 
In the second setup, 'ours with masking,' we use our three models: the landmark prediction model ($T_L$), the face editing model ($G_E$), and the lip adaptation model ($G_L$). 
The objectives of these models remain the same as in our original approach. 
However, in both the face editing ($T_L$) and lip adaptation ($G_L$) models, we mask the lower half of the input face, and therefore, we use an identity reference for both models. 
In the face editing model ($G_E$), we use a randomly selected face as the identity reference. 
In the lip adaptation model ($G_L$), we use the output of the face editing model ($G_E$) as the identity reference, which is a relatively similar approach to the identity reference used in PLGAN~\cite{yaman2024audiodriventalkingfacegeneration}. 
The third setup, 'ours (mask-free),' represents our final approach.

\begin{table}[b]
    \centering%\scriptsize
    %\vspace{-1cm}
    \setlength{\tabcolsep}{5.2pt}
    \resizebox{\linewidth}{!}{
    \begin{tabular}{l|cc|cc|cc|c}
    Method & \multicolumn{2}{c|}{\textbf{LRS2}} & \multicolumn{2}{c|}{\textbf{HDTF}} & \multicolumn{2}{c|}{\textbf{LRS2-c}} & \# Params \\
     &  IFC & LPIPS  & IFC & LPIPS  & IFC & LPIPS &  \\
    \midrule
    Wav2Lip & 0.21 & 2.5 & 0.25 & 2.8 & 0.22 & 2.4 & 36M \\
    DINet   & 0.25 & 2.5 & 0.23 & 2.5 & 0.28 & 2.5 & 139M \\
    VRT     & 0.22 & 2.5 & 0.29 & 2.8 & 0.24 & 2.5 & 181M \\
    TalkLip & 0.24 & 1.9 & 0.31 & 2.3 & 0.30 & 2.0 & 138M \\
    IPLAP   & 0.20 & 2.1 & 0.26 & 2.5 & 0.22 & 2.3 & 53M \\
    AVTFG   & 0.19 & 2.5 & 0.21 & 2.7 & 0.22 & 2.5 & 52M \\
    PLGAN   & 0.16 & 2.6 & 0.20 & 2.7 & 0.18 & 2.6 & 72M \\
    Diff2Lip & 0.15 & 2.2 & 0.25 & 2.6 & 0.16 & 2.3 & 102M \\
    Ours    & 0.15 & 2.1 & 0.20 & 2.4 & 0.17 & 2.2 & 79M \\
    \bottomrule
    \end{tabular}}
    \caption{Temporal coherence analyses using Inter frame consistency (IFC) and LPIPS. Lower is better in both metrics. LRS2-c indicates the cross-match scenario (same set with Table 2 in the main paper).}
    \label{tab:param}
\end{table}

\begin{figure}
    \centering
    \includegraphics[width=1\linewidth]{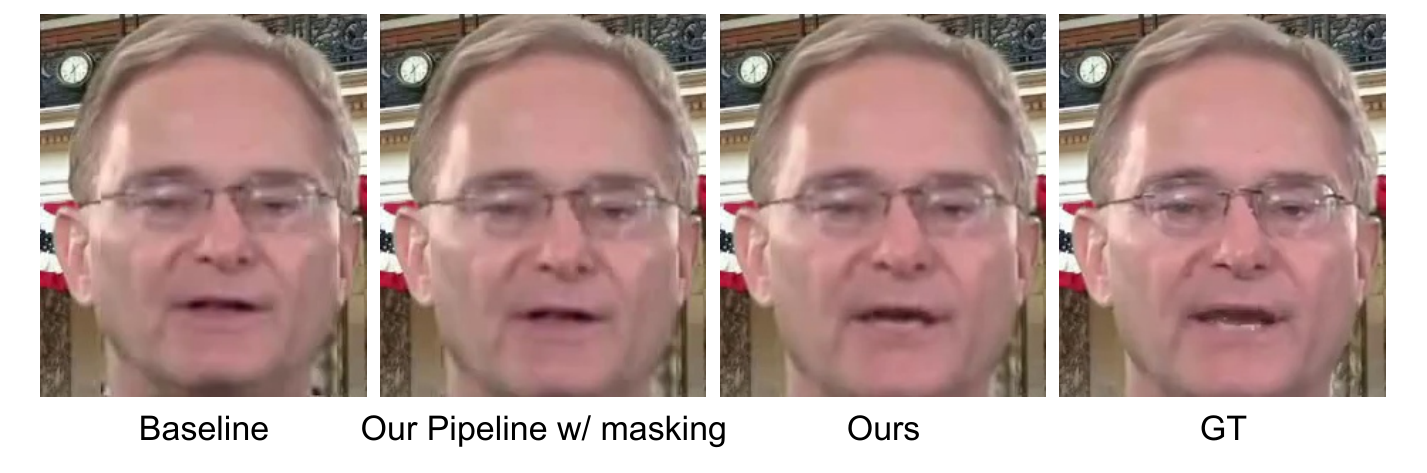}
    \caption{Generated faces with baseline, our pipeline with masking strategy, and our mask-free pipeline.}
    \label{fig:ablation_study_masking}
\end{figure}

\subsection{Hyperparameters Selection}

\noindent\textbf{Temporal dimension - T} 
Due to the extensive ablation study conducted in Wav2Lip~\cite{prajwal2020lip}, almost all works in the literature choose $T=5$. 
Therefore, we follow the literature and select $T=5$ as well in landmark adaptation model as the temporal consistency for speech is crucial.
However, in the landmark prediction model ($T_L$) and the face editing model ($G_E)$, we empirically choose $T=1$.
This is because these two models are responsible for generating faces with the neutral mouth, and as a result, there is no need for inter-frame consistency in lip movements, unlike in talking face generation.
According to our experiments, selecting different values of $T$ does not improve performance, despite slightly increased running time.

\noindent\textbf{Number of previous frames - k}
We conduct ablation study for empirically choosing $k$. 
We present the results in Table \ref{tab:hyperparameter_k}. 
According to the scores, the best performance is obtained with $k=1$.

\begin{table}[b]
    \centering%\scriptsize
    %\vspace{-1cm}
    \setlength{\tabcolsep}{5.2pt}
    %\resizebox{\linewidth}{!}{
    \begin{tabular}{l|ccc}
    $k$ &  $LD_{full}$  & $LD_{lip}$ & $LD_c$ \\
    \midrule
    $k = 1$  & 9.418 & \textbf{2.459} & \textbf{0.293} \\
    $k = 2$  & \textbf{9.419} & 2.458 & \textbf{0.293} \\
    $k = 5$  & 9.405 & 2.458 & 0.291 \\
    $k = 10$ & 9.406 & 2.455 & 0.292 \\
    \bottomrule
    \end{tabular}
    \caption{Ablation study for the hyperparameter $k$, which indicates the number of previous frames used in the landmark prediction model ($T_L$).}
    \label{tab:hyperparameter_k}
\end{table}

\noindent\textbf{Margin for face cropping after face detection}
Since the face detection model used detects faces with a tight crop, we decided to apply a margin to better cover the boundaries of the face. 
Without this margin, we sometimes slightly lose the bottom of the chin and the face boundaries. 
Our observations show that a $10\%$ margin is the most reasonable choice. 
When we use smaller margins, we still lose some face information. 
On the other hand, using more than $10\%$ introduces redundant background information unnecessarily.

\noindent\textbf{Audio parameters}
For these hyperparameters (audio frequency, window size, hop size), we follow the literature, as they have already been extensively ablated, and these selected values are considered the gold standard in audio-driven talking face generation.

\section{Additional Results}
\label{app:additional_results}

Please note that Diff2Lip~\cite{mukhopadhyay2024diff2lip} is trained on the VoxCeleb2 dataset, which consists of over 1 million face-cropped YouTube videos from more than 6,000 identities. 
This is a considerably large-scale dataset, especially when compared to LRS2, which contains only 29 hours of training data.

We demonstrate additional results in the following figures from our model.
The results show the accuracy of our whole model as well as each submodule.

\begin{figure*}
    \centering
    \includegraphics[trim={1cm 1.4cm 1cm 1.6cm},clip,width=0.62\textwidth]{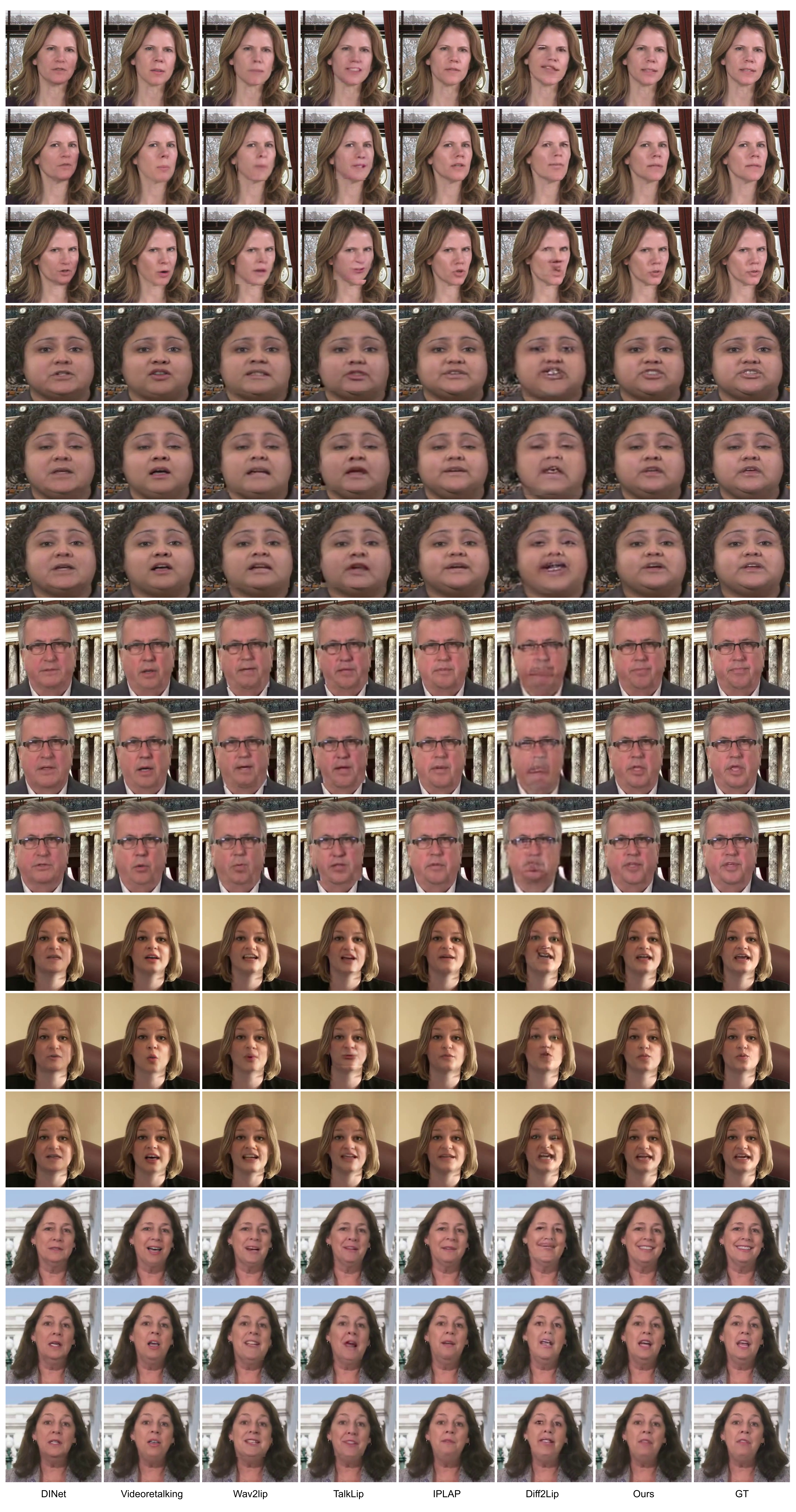}
    \caption{Additional results from the HDTF dataset. We compare the performance of different models with our model. Each set of three rows consists of sequential frames from a different video, presented in a temporally ordered way.}
    \label{fig:appendix_comparison}
\end{figure*}

In Figure \ref{fig:appendix_comparison}, we further compare our model with the existing SOTA models.

\begin{figure*}
    \centering
    \includegraphics[trim={0.8cm 1cm 0.8cm 1cm},clip,width=0.70\textwidth]{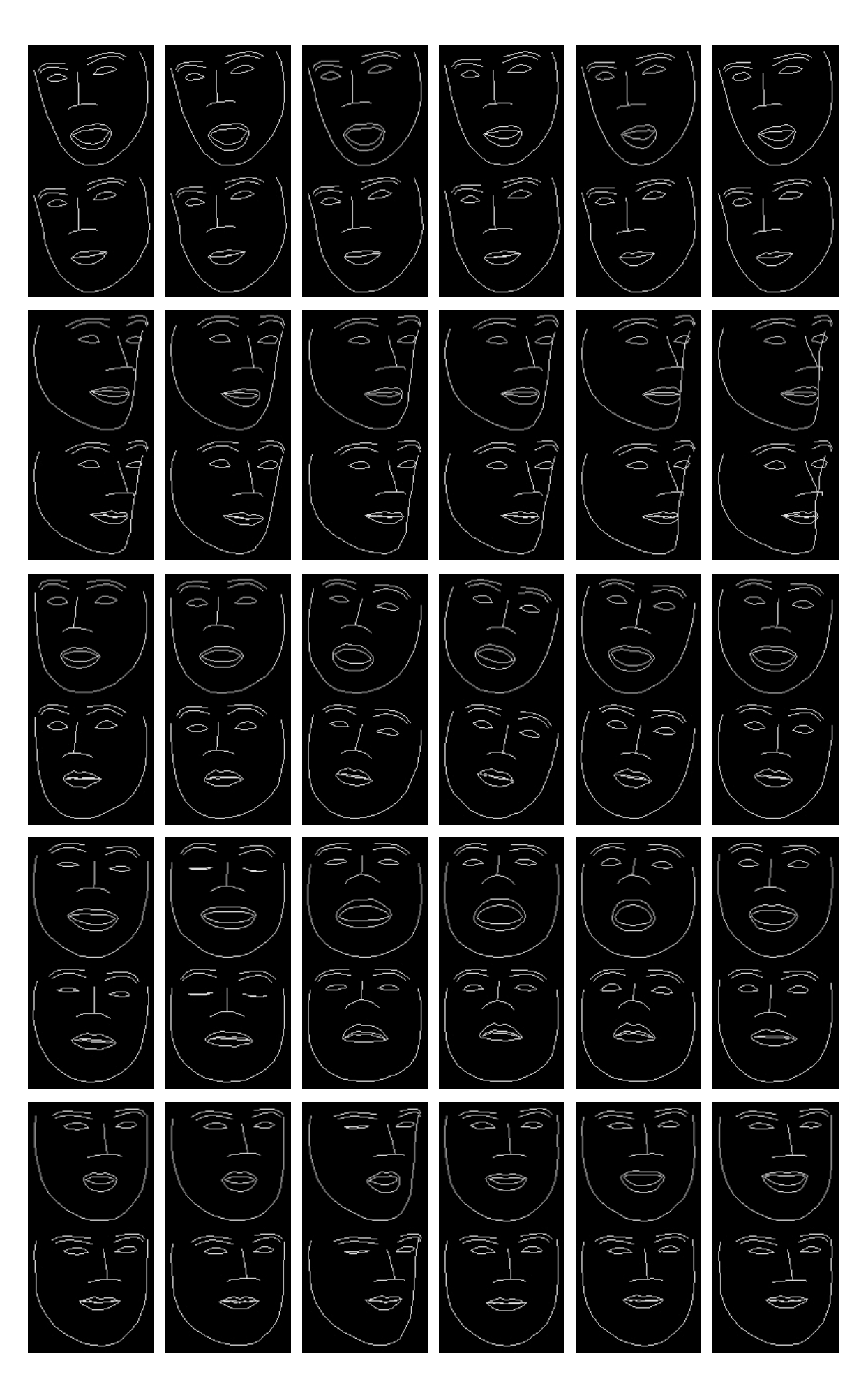}
    \caption{Generated landmark samples. In each block, the first row shows the landmark map of an input video frame, generally a talking face. The second row demonstrates the landmarks with a neutral mouth that are predicted by our landmark prediction model ($T_L$). Please note that $T_L$ predicts only the landmark vector. We visualize these landmarks to illustrate their appearance in this figure and to use them as a condition in the face editing model ($G_E$).}
    \label{fig:appendix_landmark}
\end{figure*}

In Figure \ref{fig:appendix_landmark}, we visualize the input and output of our landmark prediction model, $T_L$. 
While the first rows in each block represent the input face landmarks, the second rows depict the predicted landmark maps that have neutral mouth, which is the output of $T_L$.

\begin{figure*}
    \centering
    \includegraphics[trim={0.8cm 1cm 0.8cm 1cm},clip,width=0.64\textwidth]{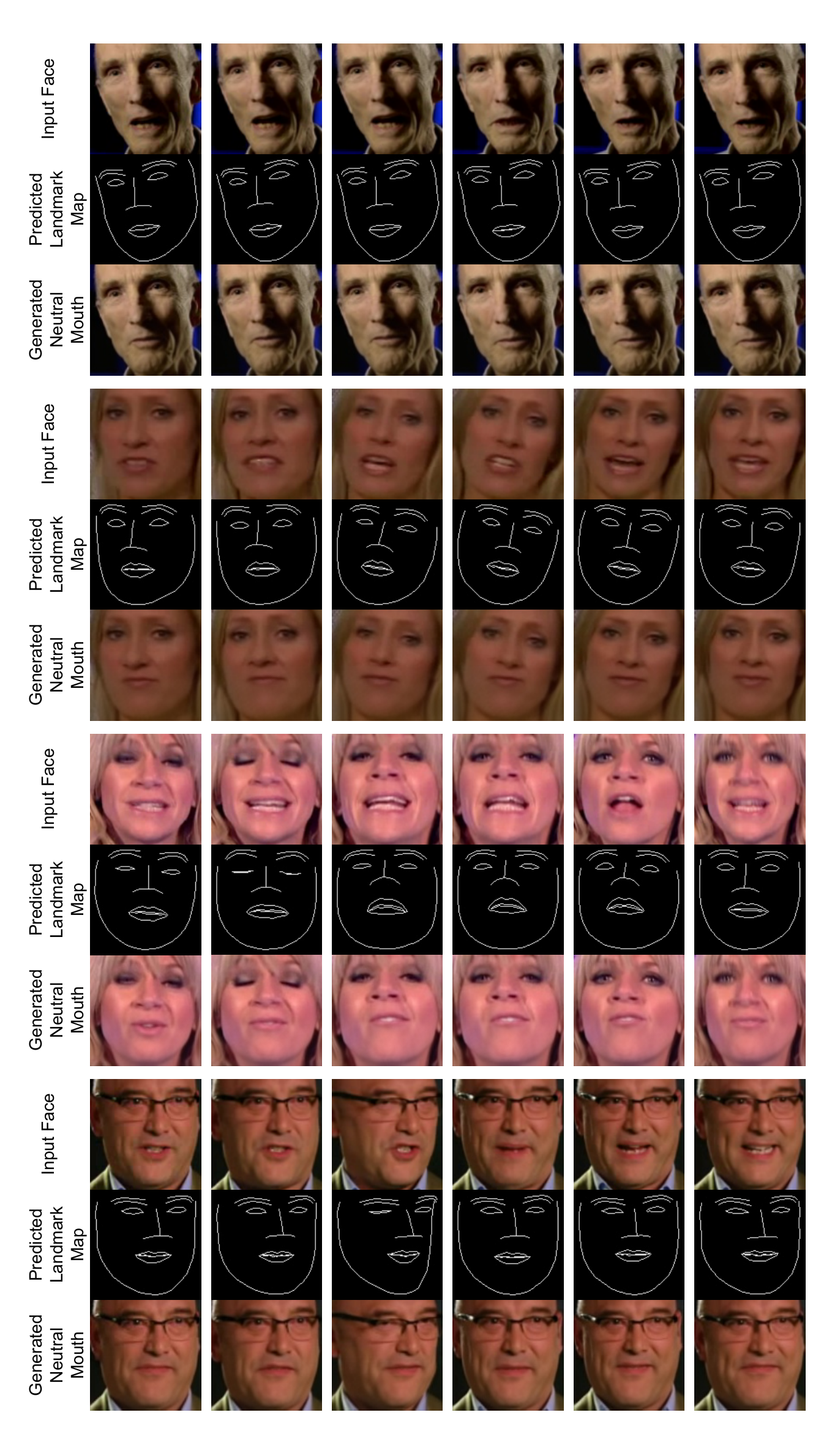}
    \caption{Output images demonstrating the performance of our face editing model ($G_E$). In each block, the first row contains the original input faces, while the second row shows the map of the predicted landmarks with a neutral mouth (visualized output of $T_L$). The last row presents the output images generated by our face editing model ($G_E$). }
    \label{fig:appendix_face_editing}
\end{figure*}

In Figure \ref{fig:appendix_face_editing}, we show the input face and input landmark map, that is predicted by $T_L$, for our face editing model, $G_E$.
We also demonsttrate the generated neutral mouth which is the outout of $G_E$.
It is the version of the input face with a neutral/closed mouth.
As can be seen from Figure \ref{fig:appendix_face_editing}, $G_E$ takes the input face and the predicted landmark map as a condition to generate a version of the input face with a neutral mouth while preserving all other details. 
The results demonstrate that $G_E$ effectively closes the mouth, while maintaining the overall facial consistency, identity, and illumination.

\begin{figure*}
    \centering
    \includegraphics[trim={0.8cm 1cm 0.8cm 1cm},clip,width=0.66\textwidth]{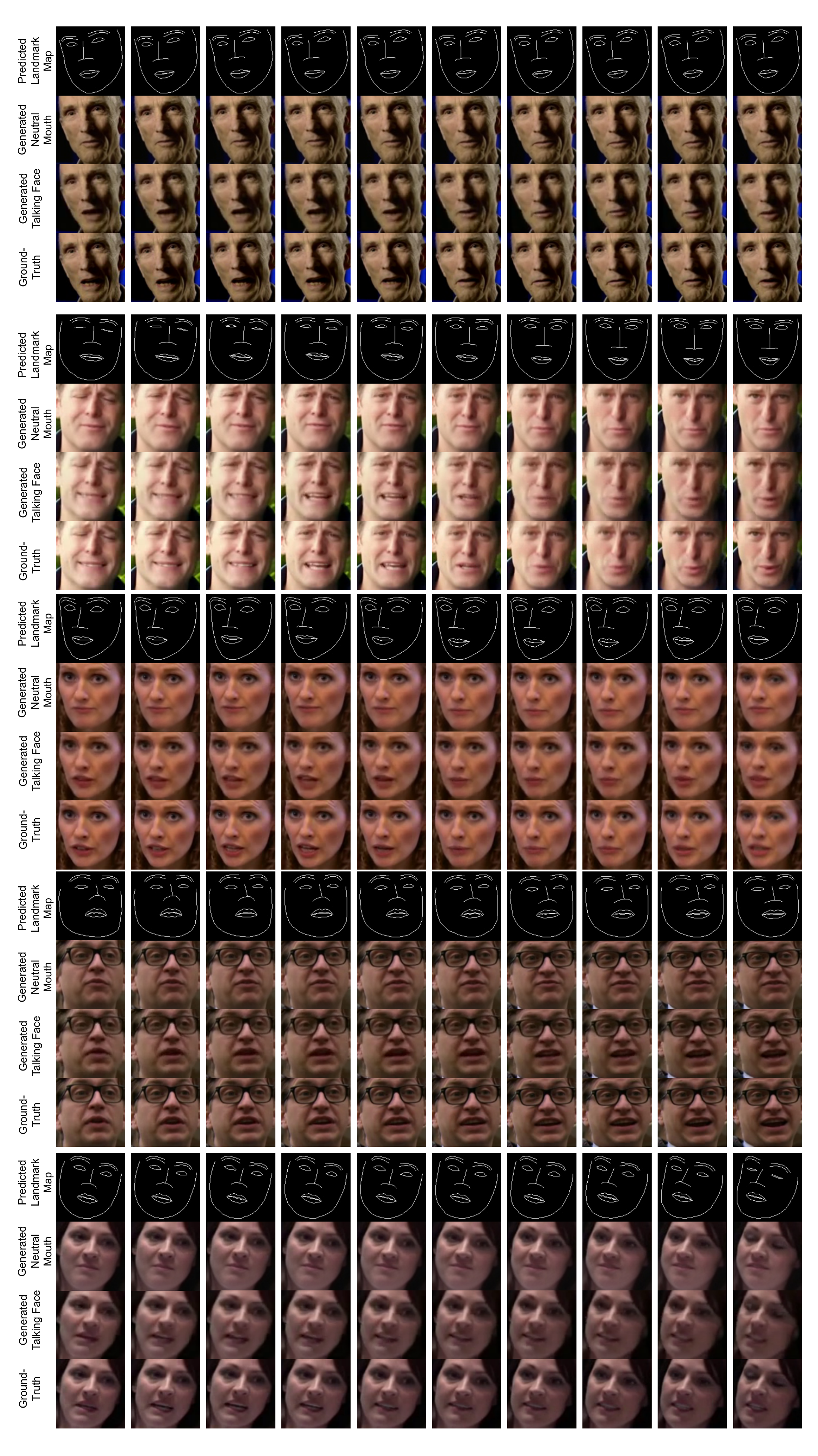}
    \caption{Demonstration of the output of each submodules along with ground-truth (GT) samples. In each block, rows demonstrate the predicted landmark map (visualization of predicted landmarks by $T_L$), generated neutral mouth (generated by $G_E$), generated talking face (generated by $G_L$), and GT face.}
    \label{fig:appendix_whole}
\end{figure*}

In Figure \ref{fig:appendix_whole}, we present the outputs of each submodule: $T_L$, $G_E$, $G_L$.
In each block:
\begin{itemize}
    \item The first row visualizes the predicted landmark map (output of $T_L$).
    \item The second row shows the output of $G_E$, which is a face image with a neutral mouth. During this process, $G_E$ uses the predicted landmark map as a conditioning input.
    \item  The third row has the output of $G_L$, which is the generated talking face conditioned on the audio and the neutral-mouth image (generated neutral mouth), generated by $G_E$.
    \item The last row contains the ground-truth (GT) face images.
\end{itemize}

Each block consists of ten sequential frames from different randomly selected videos in the HDTF dataset.

\clearpage

\clearpage

\clearpage

\clearpage

\end{document}